# Zhinv: Real-time hub-height wind field reconstruction using only local sparse observations

Zongwei Zhang[1,2,3], Chin Chun Ooi[2], Lianlei Lin[1,3,*], Sheng Gao[1,3], Tiantian He[2], Yew Soon Ong[2,4], Junkai Wang[1,3], Hangyi Yu[1,3], Jiaqi Zhang[5], Hanqing Zhao[1,3], Yu Zhang[1,3]

[1] School of Electronic and Information Engineering, Harbin Institute of Technology, Harbin, 150001, Heilongjiang, China

[2] Centre for Frontier AI Research, Agency for Science, Technology and Research (A*STAR), Singapore 138632, Singapore

[3] Technological Innovation Center of Littoral Test, China

[4] College of Computing & Data Science, Nanyang Technological University, 639798, Singapore

[5] College of Intelligent Systems Science and Engineering, Harbin Engineering University, Harbin, 150001, Heilongjiang, China

* Corresponding author: Lianlei Lin (linlianlei@hit.edu.cn)

**Abstract:** The high proportion of wind power connected to the grid places higher demands on fine-grained knowledge of regional wind fields. Since the wind information directly obtainable in actual operations is mostly sparse, discrete, and irregularly distributed local observations, it is difficult to directly meet the needs of tasks such as wind power regulation, wind resource assessment, and low-altitude environmental perception of continuous regional wind fields. Therefore, we propose Zhinv, an end-to-end reconstruction framework that directly weaves sparse and irregular observations into a fine-grid wind field at hub-height. Experiments in Northeast China, Europe, and Southeast Asia demonstrate that Zhinv can accurately, robustly, and efficiently reconstruct fine-grid wind fields from sparse observations, reducing the error by about 66% compared with Kriging. With local wind-power observations as input, Zhinv enables wind power centers to bypass NWP and complex assimilation processes, supporting direct and real-time wind resource assessment from locally available data.

# 1 Introduction

With the ongoing global energy transition, large-scale wind power grid connection has become an important pillar for achieving carbon neutrality in modern power systems [1,2,3]. However, wind energy exhibits significant spatiotemporal volatility, posing a serious challenge to real-time power balance and power flow security. To effectively reduce wind curtailment rates and optimize reserve capacity, the power grid dispatch center urgently needs spatial distribution of wind fields in wide-area wind power clusters [4-5]. However, in actual operations, most of the wind information that can be directly obtained comes from previously deployed meteorological stations, anemometer towers, lidar, and distributed sensor networks. These observation data are usually sparse, discrete, and irregularly distributed in space, a contrast from the continuous and regular wind fields required by downstream applications [6]. Therefore, recovery of regional wind fields using limited discrete observations has become a key issue in the application of wind power meteorological information [5].

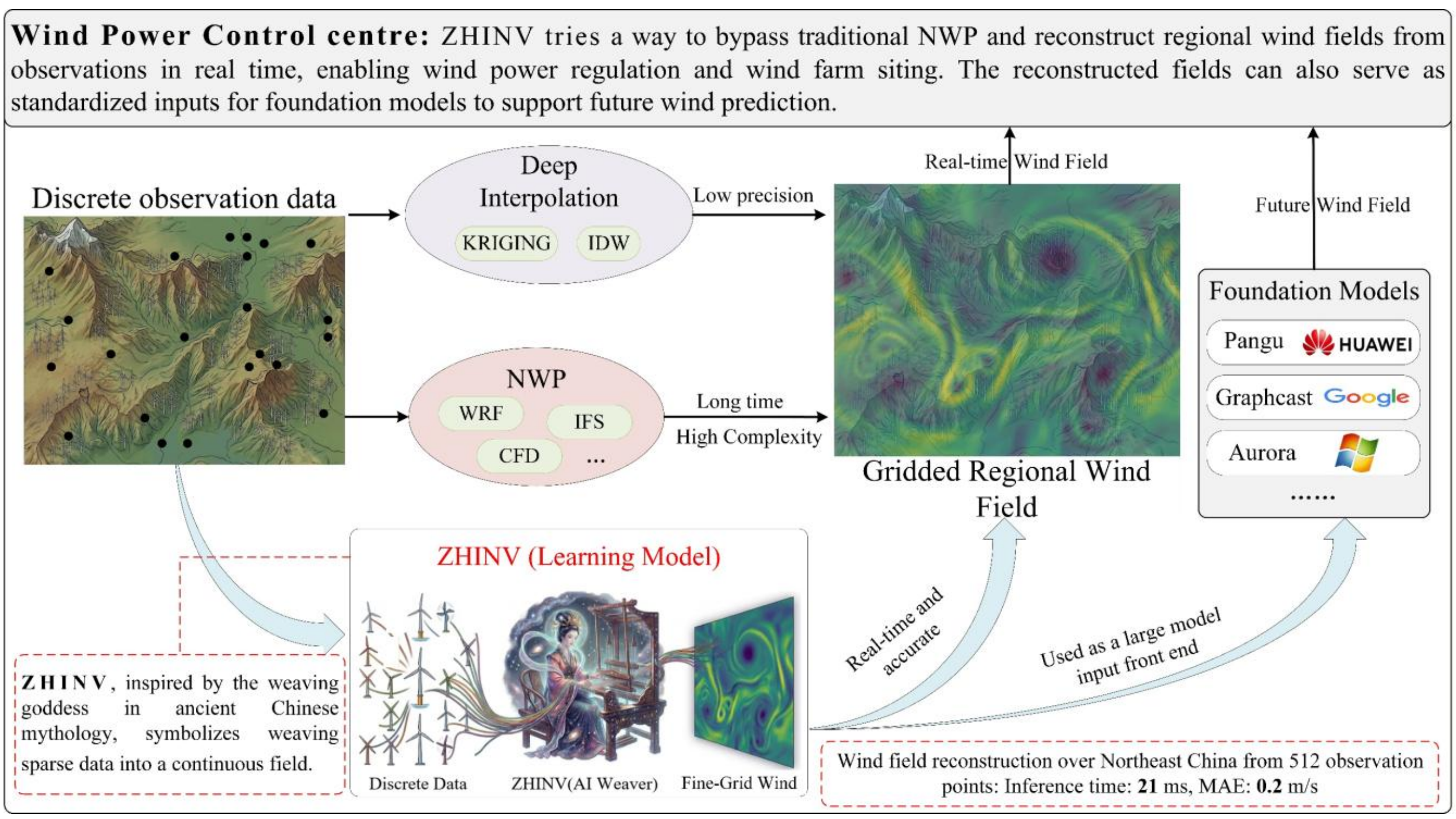

**Figure 1.** Overview of Zhinv: weaving sparse observations into a regional wind field.

Existing methods for constructing regional wind fields based on discrete observations mainly comprise three categories: physical solutions [7-9], spatial interpolation [10-11], and data-driven methods [12-14]. Numerical weather prediction (NWP) can generate spatiotemporally continuous and physically consistent wind fields based on physical equations and has long been an important foundation for wind environment analysis [7,15]. It can utilize discrete observation information when obtaining regional wind fields through processes such as assimilation, constraint and downscaling [7,8]. However, this path relies on knowledge of complex initial and boundary conditions, and costly assimilation processes, numerical solutions and post-processing links, making it difficult to simultaneously consider timeliness and deployment flexibility in local wind power scenarios [16,17]. Similarly, computational fluid dynamics (CFD) methods can characterize local flow structures under complex terrain with higher spatial resolution, and have important value in wind resource assessment, wake analysis and fine wind environment simulation [18-20]. However, CFD also relies on fine mesh generation, strict boundary settings and high computational cost, so it is more suitable for high-fidelity local simulation and cannot meet the needs of rapid reconstruction and real-time update of regional wind fields in wind power scenarios [21].

In contrast, spatial interpolation methods such as inverse distance weighting [22] and Kriging [23] can directly utilize discrete observations to quickly estimate wind speeds at unobserved locations at a lower cost, thus possessing strong practicality in local wind information completion [10]. However, these methods are mainly based on assumptions such as proximity, stationarity, or local smoothness, and can perform poorly when reconstructing the true spatial organization of complex wind fields [24,25]. When wind fields are simultaneously affected by heterogeneous underlying surfaces, and the interaction of local flows, relying solely on distance decay or statistical correlation makes it difficult to accurately restore their spatial structure and propagation characteristics [24,26]. Data-driven methods provide new insights into this problem. One type of method, as exemplified by DeepKriging [23] and KCN [27], enhances traditional interpolation by learning more complex spatial correlations. Another type of method draws on the idea of image completion to reconstruct the complete field from partial observations [28,29]. These studies show that deep learning models can outperform classical interpolation methods in the recovery of dense fields [5,28]. However, most existing methods still focus on numerical completion or maintaining the continuity of local structures and usually lack explicit modeling of the influence of terrain and the wind field propagation mechanism. Therefore, in complex environments, they find it difficult to recover the regional wind field with realistic structure and physical rationality [12,30].

Meanwhile, regular grid-driven meteorological models [31,32,33,34], such as Pangu [16], GraphCast [31], and FourCastNet [32], have shown strong spatiotemporal representation capabilities in large-scale meteorological field prediction. However, the input of such models is usually pre-set as a structured grid field, so they cannot solve the prerequisite problem of reconstructing a regional wind field from sparse discrete observations [35,36]. Graph models oriented towards station networks or sensor networks can naturally handle irregular discrete inputs and have been used for tasks such as wind speed prediction [37-39]. However, their output usually remains at the point level, making it difficult to further form a gridded and complete regional wind field expression [37,40,41].

Existing research, including models designed for regular-grid fields or point-level outputs, has significantly improved local completion capability and spatiotemporal prediction capability. However, a unified and effective framework for constructing structurally sound regional wind fields from discrete observations is still lacking. In practical operational systems, readily available data typically consist of node-level, irregularly distributed local observations, while downstream prediction, analysis, and applications rely on structurally complete regional wind fields [35,42]. Therefore, the core of this problem is not merely improving the accuracy of a single model but bridging the gap between these two representations by stably transforming discrete node information into field-level expressions. With such a bridge, the model can serve as a back-end module that converts discrete information into structured field representations, and also as a front-end module that provides structured inputs at the current time for advanced meteorological foundation models. Recent studies have also treated global field reconstruction driven by sparse sensor observations as a general field reconstruction problem and have achieved the recovery of complex spatial fields from a small number of observations through attention mechanisms [43]. However, these methods are typically geared towards general field reconstruction tasks and do not specifically model topographic influence and propagation mechanisms in wind field reconstruction.

To address this issue and achieve rapid reconstruction of hub-height wind fields, we propose Zhinv, as shown in Figure 1. This method is an end-to-end wind field reconstruction framework for sparse and irregular observations, capable of directly generating high-fidelity fine-grid regional wind fields from discrete nodes. Experimental results show that Zhinv outperforms existing baseline methods in both spatial structure preservation and overall error control, demonstrating stronger wind field recovery capabilities. More importantly, Zhinv's value lies not only in high-precision wind field reconstruction but in providing a general spatial representation conversion mechanism. On the one hand, as a back-end field representation module, Zhinv can further recover regional wind fields from discrete node predictions, thereby supporting more intuitive wind field situation analysis and downstream wind power assessment. On the other hand, as a front-end field construction module, Zhinv can quickly convert sparse observations into gridded wind fields, providing high-quality initial fields for prediction models that require regular grid inputs, thus offering a new approach to traditional NWP assimilation and intermediate processing procedures.

# 2 Methodology

Figure 2 illustrates the overall architecture of Zhinv. This model is designed to reconstruct regular grid wind fields from discrete observation points under complex terrain conditions. It aims to overcome the limitations of traditional NWP processes, which are complex and have high inference overhead, and achieve rapid overall reconstruction of regional wind fields. As shown in Figure 2, the network input mainly consists of two parts: first, discrete observation point data, including u/v-wind speed $X_{wind} \in R^{2\times N}$, spatial coordinates $X_{coord} \in R^{2\times N}$, and point-level terrain features $X_{terrain} \in R^{4\times N}$; second, regular grid terrain data $F_{Terrain} \in R^{4\times H\times W}$, containing physical information such as elevation, slope, aspect, and surface roughness of the target area. To address the uneven interactions between observation points, Zhinv breaks through the limitations of pure geometric distance by using a point-wise graph encoder, which incorporates physical terrain factors as modulation conditions into the graph topology to extract spatial features of discrete points under complex terrain. Subsequently, a terrain-distance constrained cross-attention Interpolator is used to explicitly introduce grid terrain as a physical constraint, integrating spatial distance and terrain differences to map and transform discrete node features into physically consistent regular grid features. Finally, a multi-scale wind field refinement decoder uses global terrain priors to guide the decoding of grid spatial features and generate refined details in wind speed evolution, ultimately completing the overall reconstruction of the wind field from discrete points to a regular grid.

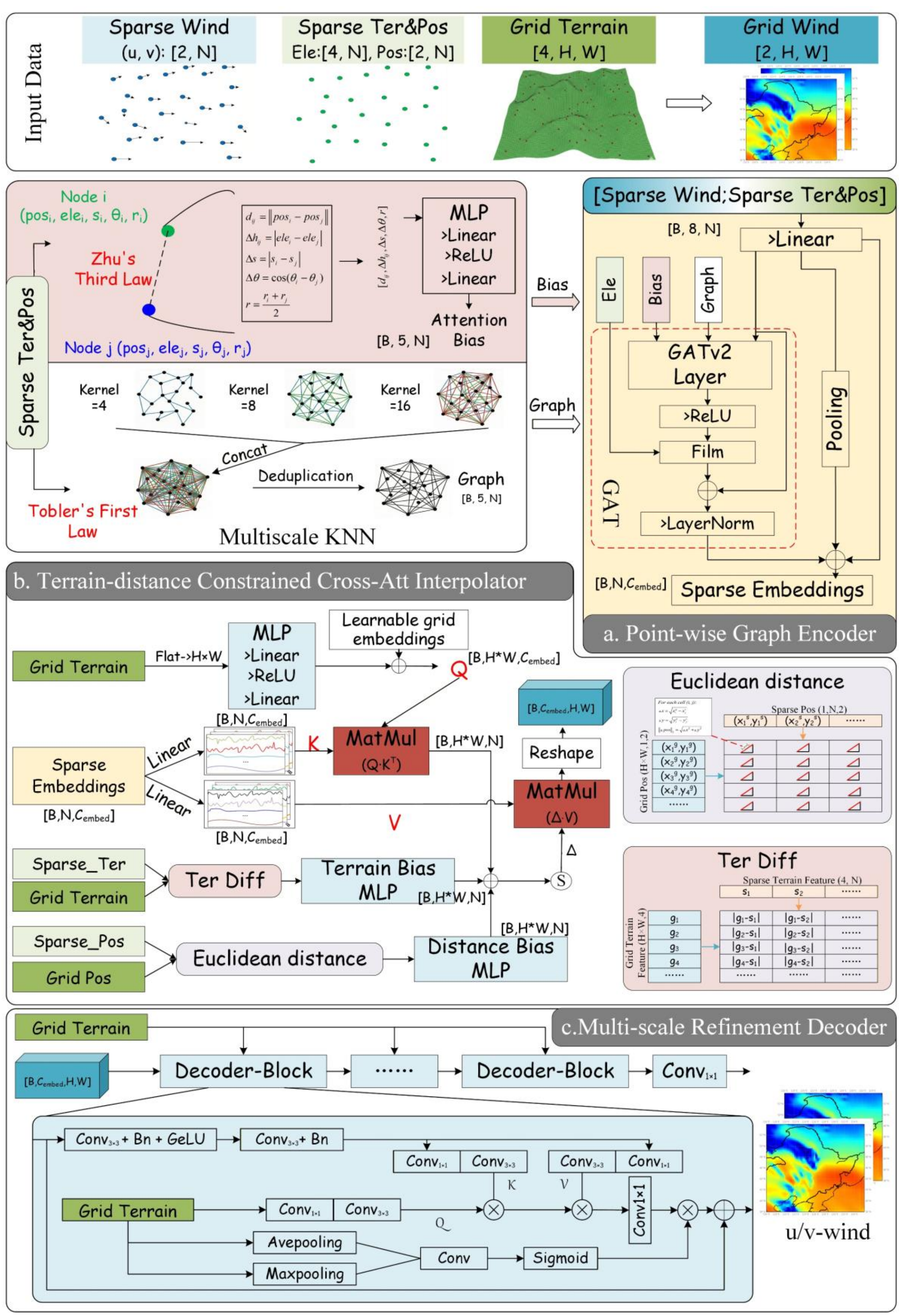

**Figure 2.** Overall architecture of Zhinv. Given sparse wind speed, sparse terrain and positional information, and regular-grid terrain data, the model reconstructs a dense regular wind field through a) a point-level graph encoder, b) a terrain-distance constrained cross-attention interpolator, and c) a multi-scale enhancement decoder.

## 2.1 Point-wise graph encoder

To extract sparse representations that reflect local wind field propagation relationships from spatially irregular discrete observation points, we designed a point-level graph encoder, as shown in Figure 2 (a). This module is motivated by the idea that spatial proximity determines the strength of local correlations (the First Law of Geography [44]), while topographic heterogeneity further modulates information propagation between points (the Third Law of Geography [45]). Specifically, a multi-scale adjacency graph is first constructed based on the coordinates of discrete observation points, and edge-level attention biases are generated by incorporating topographic attributes. Subsequently, discrete wind speed, topographic information, and positional information are jointly used as node inputs, and a deep graph attention network [46] is employed to achieve local-global joint encoding, thereby providing terrain-constrained sparse feature representations for the subsequent continuous point-to-grid mapping.

The input for discrete wind speed observation is $X_{wind} \in R^{2\times N}$, where $N$ represents the number of discrete observation points, and the two channels correspond to the two components of the wind field. Discrete topographic features are represented by $X_{terrain} = [z, s, \theta, \rho]^T \in R^{4\times N}$, where $z$, $s$, $\theta$, and $\rho$ represent elevation, slope, aspect, and surface roughness, respectively. Considering the periodicity of the aspect variable, we will subsequently expand it into sine and cosine forms to avoid discontinuities caused by abrupt angle changes. Discrete point locations are represented by $X_{coord} \in R^{2\times N}$.

First, we construct a multi-scale KNN graph based on the locations of discrete points. For any two observation nodes $i$ and $j$, their relative geometric distance in coordinate space is defined as:

$$d_{(i,j)} = \left\| x_i^{coord} - x_j^{coord} \right\|_2$$

To enable the encoder to simultaneously capture local interactions at different spatial radii, we set up a multi-scale neighborhood set $\mathcal{K} = \{4, 8, 16\}$ and construct corresponding KNN graphs for each. After concatenating and de-duplicating the edge sets at each scale, we obtain the final graph connectivity $A_{edge} \in R^{2\times E}$, where $E$ is the final number of edges.

After determining the graph structure, we construct edge-level attention biases based on topographic differences between the nodes at both ends of an edge to apply the Third Law of Geography [45]. For any edge $e$ ($r_e$ and $c_e$ represents the index of the edge's start point and the index of its end point), we define five types of physical edge characteristics: geometric distance, elevation difference, slope difference, aspect consistency, and path average roughness:

$$f_e^{(1)} = \left| x_{r_e}^{coord} - x_{c_e}^{coord} \right|_2$$
$$f_e^{(2)} = \left| z_{r_e} - z_{c_e} \right|$$
$$f_e^{(3)} = \left| s_{r_e} - s_{c_e} \right|$$
$$f_e^{(4)} = \cos\left( \theta_{r_e} - \theta_{c_e} \right)$$
$$f_e^{(5)} = \frac{\rho_{r_e} + \rho_{c_e}}{2}$$

After concatenation, we obtain the feature vector $f_e^{edge} = \left[ f_e^{(1)}, f_e^{(2)}, f_e^{(3)}, f_e^{(4)}, f_e^{(5)} \right] \in R^5$ of edge $e$, and thus the set $F^{edge} \in R^{E\times5}$ of edges $E$. Subsequently, we map this set to edge-level attention biases using a multilayer perceptron:

$$Bias = \mathrm{MLP}_{\mathrm{edge}}\left( F^{edge} \right)$$

where $Bias \in R^E$ represents the attention bias vector for all edges. This bias term is added to the subsequent graph attention calculation as an explicit terrain prior, so that the propagation of information between points depends not only on feature correlation but also on the modulation of complex terrain conditions.

Under complex terrain conditions, the distribution of wind fields is not determined solely by the observations themselves, but is jointly influenced by terrain morphology, underlying surface resistance, and spatial positional relationships. To ensure that node representations incorporate local dynamical states and environmental constraint information, which is crucial to the model's ability to characterize the spatial heterogeneity of wind fields under complex underlying surface conditions, we fuse elevation, terrain, and other relevant information with discrete wind speed and use them jointly as inputs to the encoding network:

$$X_{input} = \left[ X_{wind}, z, s, \sin(\theta), \cos(\theta), \rho, X_{coord} \right] \in R^{9\times N}$$

Node features are obtained by projecting the node input:

$$H^{(0)} = \mathrm{Linear}\left( X_{input}{}^{T} \right) \in R^{N\times C}$$

where $C$ is the embedding dimension and $\mathrm{Linear}(\cdot)$ represents the linear projection operation. Subsequently, the node features are fed into a multi-layer GAT Block [46] for local relation modeling. Let $\mathrm{GAT}^{(l)}(\cdot)$ be the graph attention aggregation operation in layer $l$, which internally performs neighborhood attention calculation and feature aggregation based on edge connectivity $A_{edge}$ and terrain bias $Bias$. The local aggregation result of the discrete node features can then be expressed as:

$$M^{(l)} = \mathrm{GAT}^{(l)}(H^{(l)}, A_{edge}, Bias)$$

To further enhance the model's adaptability to complex underlying surface variations, we introduce terrain-condition-based FiLM modulation after each layer's graph attention. We select node elevation and roughness to form a condition vector $T_{cond} = [z, \rho] \in R^{N\times2}$, from which scaling factor $\gamma$ and bias term $\beta$ are generated:

$$\gamma = \mathbf{W}_{\gamma} T_{cond} + \mathbf{b}_{\gamma}$$
$$\beta = \mathbf{W}_{\beta} T_{cond} + \mathbf{b}_{\beta}$$

Therefore, the feature update after nonlinear activation and conditional modulation is as follows:

$$\hat{M}^{(l)} = \gamma \odot \mathrm{ReLU}(M^{(l)}) + \beta$$

where, $\odot$ represents element-wise multiplication. Then, the aggregated feature $\hat{M}^{(l)}$ and the node feature $h_i^{(l)}$ residuals are added together and normalized by layers to obtain the updated node representation $h_i^{(l+1)}$. By stacking $L$ layers of graph attention modules, the local branch output $H_{local} \in R^{N \times C}$ can be obtained.

In addition to the local propagation branch, we further construct a global context branch, supplementing the overall background information at the region scale through global max pooling, to obtain the global branch output $H_{global} \in R^{N \times C}$. Finally, we perform a weighted fusion of the local and global branches and form a residual connection with the initial projected features to obtain the final output of the point-level encoder:

$$H_{enc} = H^{(0)} + \alpha H_{local} + \beta H_{global}$$

where $H_{enc} \in R^{N \times C}$, and $\alpha$ and $\beta$ are learnable scalar parameters.

## 2.2 Terrain–distance constrained cross-attention interpolator

After obtaining the sparse feature representation $H_{enc}$ from the point-level map encoder, it needs to be mapped from the irregular discrete observation space to the desired regular grid space for continuous reconstruction of the regional wind field from points to surfaces. Therefore, we designed a terrain-distance constrained cross-attention interpolator, as shown in Figure 2 (b). This module uses the target location on the regular grid as the query (Q) and the encoded features of discrete observation points as the key (K) and value (V), establishing feature associations between grid points and discrete observation points through a cross-attention mechanism. Simultaneously, terrain bias and distance bias are explicitly introduced into the attention calculation, making the interpolation process not only dependent on feature similarity but physically consistent with the known regional terrain and spatial distance decay.

The target regular grid size is $M = H \times W$, and the corresponding regular grid terrain feature is $F_{Terrain} \in R^{4 \times H \times W}$. The two-dimensional coordinates of all target grid points are $P_{grid} \in R^{M \times 2}$. First, a learnable grid query embedding $E^q \in R^{M \times C}$ is introduced for each target location on the regular grid to provide an initial query representation for the target grid point, enabling the regular grid location to participate in cross-attention calculation as an active query endpoint, and introducing distinguishable spatial priors for different grid locations. Then, a sine and cosine expansion is performed on the slope aspect to obtain the grid-side terrain representation as follows:

$$\tilde{F}_{Terrain} = [Z_{grid}, S_{grid}, sin(\theta_{grid}), cos(\theta_{grid}), \rho_{grid}] \in R^{5 \times H \times W}$$

Subsequently, the regular grid terrain features are unfolded and mapped into query terms using a multilayer

perceptron:

$$E^{t} = \mathrm{MLP}_{q}(\mathrm{Flatten}(\tilde{F}_{Terrain}^{\ \ T})) \in R^{M \times C}$$

Therefore, the query vector (Q) for the target grid point can be represented as:

$$Q = \mathrm{Linear}_{q}(E^{q}) + E^{t}$$

For the coded feature $H_{enc}$ of discrete observation points, the key (K) and value (V) are obtained through linear mapping:

$$K = \mathrm{Linear}_{\mathrm{k}}(H_{enc})$$
$$V = \mathrm{Linear}_{\mathrm{v}}(H_{enc})$$

Therefore, we can first obtain the basic attention score based on feature similarity:

$$A_{mn}^{Base} = \frac{Q_{m} K_{n}^{T}}{\sqrt{C}}$$

where $A_{mn}^{Base}$ represents the basic correlation strength between target grid point $m$ and discrete observation point $n$.

To guide the attention mechanism to more accurately incorporate the influence of other discrete observation points on a target grid location, we explicitly construct two prior correction terms, namely terrain bias and distance bias, on top of the basic attention score. In this way, the interpolation weights are determined not only by feature similarity, but also jointly modulated by local terrain consistency and the spatial decay pattern. For the terrain bias term, let the terrain attributes of the grid point $m$ and the discrete observation point $n$ be denoted by $(z_{m}, s_{m}, \theta_{m}, \rho_{m})$ and $(z_{n}, s_{n}, \theta_{n}, \rho_{n})$, respectively. Then, the terrain differences between them, including the absolute elevation difference, the absolute slope difference, the absolute roughness difference, and aspect consistency, can be expressed as:

$$\Delta z_{mn} = | z_{m} - z_{n} |$$
$$\Delta s_{mn} = | s_{m} - s_{n} |$$
$$\Delta \rho_{mn} = | \rho_{m} - \rho_{n} |$$
$$\Gamma_{mn}^{\theta} = \cos(\theta_{m} - \theta_{n})$$

The above four physical differences are concatenated into a terrain relation vector and mapped through a nonlinear transformation to generate the terrain bias matrix $B_{terrain} \in R^{M \times N}$. This bias term ensures that when there are significant terrain differences between a target grid point and a discrete observation point, their mutual influence is adaptively suppressed even if they are spatially close. For example, when the target grid point and the observation point are located on opposite sides of a ridge and a valley bottom, respectively, geometric proximity alone is insufficient to reflect their true wind field propagation relationship, whereas the terrain bias can correct for this effect.

For the distance bias term, the Euclidean distance between the coordinates $p_m^{grid} \in R^2$ of grid point $m$ and the coordinates $x_n^{coord} \in R^2$ of discrete observation point $n$ can be obtained as follows:

$$d_{(m,n)} = \left\| p_m^{grid} - x_n^{coord} \right\|_2$$

Based on this, a distance matrix A can be constructed, which is further fed into a distance mapping network to generate the distance bias matrix B. This term explicitly follows the fundamental law that spatial interaction decays with distance, so that discrete observation points located farther away are generally assigned lower influence weights during interpolation, while preserving the network's ability to adaptively learn complex propagation patterns. Finally, the basic attention scores, terrain bias, and distance bias are jointly combined to obtain the normalized attention weights between target grid points and discrete observation points:

$$\alpha_{mn} = \frac{\exp\left(A_{mn}^{\text{Base}} + B_{\text{terrain}}^{(m,n)} + B_{\text{dist}}^{(m,n)}\right)}{\sum_{j=1}^{N} \exp\left(A_{mj}^{\text{Base}} + B_{\text{terrain}}^{(m,j)} + B_{\text{dist}}^{(m,j)}\right)}$$

Based on this attention weight, the latent features of the m-th target grid point can be obtained as follows:

$$h_m^g = \sum_{n=1}^{N} \alpha_{mn} V_n$$

Stacking the features of all grid points yields a latent representation $H^g \in R^{M \times C}$ of the regular grid in the feature space. Finally, this representation is reshaped into a two-dimensional grid form $F^g \in R^{C \times H \times W}$, which serves as the input to the subsequent decoder.

## 2.3 Multiscale wind field refinement decoder

Although the interpolated result $F^g \in R^{C \times H \times W}$ is already represented on a regular grid, its spatial details still mainly originate from the propagation of discrete point features, which is insufficient to fully recover local wind field details under complex terrain conditions, especially in regions with abrupt terrain changes such as ridges, slopes, and valleys. To address this issue, we design a multi-scale wind field refinement decoder, as shown in Figure 2 (c). The decoder progressively refines grid features through multiple terrain-aware convolutional blocks and continuously guides the feature recovery process with terrain priors, thereby generating continuous wind field results that are more consistent with the effects of complex terrain surfaces.

Assuming $X^{(0)} = F^g$, then the input features of the $l$-th layer can be represented as $X^{(l-1)} \in R^{C \times H \times W}$. First, local spatial modeling is performed using two convolutional layers:

$$U^{(l)} = \text{BN}_2(\text{Conv}_{3\times3,d_l}(\phi(\text{BN}_1(\text{Conv}_{3\times3}(X^{(l-1)})))))$$

where $\text{Conv}_{3\times3}$ represents standard 3×3 convolution, $\text{Conv}_{3\times3,d_l}$ represents dilated convolution with an expansion rate of $d_l$, BN represents batch normalization, and $\phi$ represents the GELU activation function. Our aim with this design is to enable the decoder to progressively expand its receptive field while preserving local texture information, thus adapting to wind disturbance patterns at different scales in complex terrain.

On this basis, to explicitly introduce terrain priors, we follow Restormer [47] and construct a lightweight cross-attention mechanism using convolution, in which terrain serves as the query, while wind field features serve as the keys and values. In this way, terrain conditions actively guide feature selection and reorganization during the decoding stage, enabling wind field reconstruction at different grid locations to adaptively perceive the underlying surface environment in which they are situated. The terrain features obtained from the interpolation module are mapped into query tensors through a $1 \times 1$ convolution and a depth-wise separable convolution:

$$Q^{(l)} = \mathrm{DWConv}(\mathrm{Conv}_{1\times1}^{q}(\tilde{F}_{Terrain}))$$

where $\mathrm{Conv}_{1\times1}^{q}$ represents the convolution operation that maps the 5-channel terrain input to a C-dimensional feature space, and $\mathrm{DWConv}$ represents the depth-wise convolution. Simultaneously, the current wind field feature $U^{(l)}$ is mapped to keys and values:

$$K^{(l)} = \mathrm{DWConv}(\mathrm{Conv}_{1\times1}^{k}(U^{(l)}))$$
$$V^{(l)} = \mathrm{DWConv}(\mathrm{Conv}_{1\times1}^{v}(U^{(l)}))$$

Subsequently, cross-attention weights are calculated, and the tensors are weighted and aggregated to obtain the terrain-guided feature enhancement results:

$$O^{(l)} = \mathrm{Softmax}((Q^{(l)}K^{(l)T}) \text{e } \tau)V^{(l)}$$

where $\tau$ is a learnable temperature parameter used to adaptively adjust the response intensity of different attention heads. Next, the spatial attention module [48] is used to obtain the spatial weight map $\sigma \in R^{1\times H\times W}$ of the wind field features, which is used to calibrate and filter more significant local structural responses:

$$Z^{(l)} = \sigma \text{ e } \left(U^{(l)} + O^{(l)}\right)$$

To preserve the original feature information and stabilize the deep training process, we employ residual connections at the end of each terrain-aware block to obtain the output of the $l$ layer:

$$X^{(l)} = X^{(l-1)} + Z^{(l)}$$

The final wind field prediction result is obtained through 1×1 convolution:

$$Y = Conv_{1\times1}^{out}\left(X^{(L)}\right) \in R^{C_{out}\times H\times W}$$

where $C_{out} = 2$ represents the two components of the output wind field.

## 2.4 Masked frequency-physical hybrid loss

To improve wind field reconstruction during training, we develop the masked frequency-physical hybrid loss (MFPH-Loss) function. This loss guides model optimization from three complementary aspects: 1) scalar fidelity, 2) high-frequency structure recovery, and 3) physical consistency constraint. This enables the model to not only accurately recover wind speed values in unknown regions, but also preserve local details and the dynamical plausibility of the overall wind field.

Assuming that inference field is $Y_{\mathrm{SR}} = (U_{\mathrm{pred}}, V_{\mathrm{pred}})$, ground truth field is $Y_{\mathrm{GT}} = (U_{\mathrm{true}}, V_{\mathrm{true}})$. Since the wind speed values at some sparse observation points in the input are already known, these positions should not be repeatedly constrained as pixel-level supervision targets. Therefore, we introduce a binary mask $M_{Observ} \in \{0,1\}^{H \times W}$, where the positions of known observation points are set to 0 and the remaining areas to be reconstructed are set to 1. We first define the masked reconstruction loss as:

$$L_1 = \| M_{observ} \odot (Y_{\mathrm{SR}} - Y_{\mathrm{GT}}) \|_1$$

However, relying solely on L1 loss may smooth out high-frequency structures. Therefore, we include a frequency-domain loss term using Laplacian decomposition. Specifically, we apply a binomial kernel to the wind field via convolution to extract low-frequency components and then subtract these to isolate the high-frequency components:

$$Y_{\mathrm{high}} = Y - K_{\mathrm{blur}} * Y$$

where $K_{\mathrm{blur}}$ is a fixed-weight smoothing kernel. L1 loss is computed separately for the high-frequency portions of the prediction and ground truth, enhancing the model's ability to capture local variations and energy gradients:

$$L_{\mathrm{high}} = \| Y_{\mathrm{SR,high}} - Y_{\mathrm{GT,high}} \|_1$$

Furthermore, to ensure that the generated wind field meets physical constraints, mainly divergence-free and vorticity consistency [49], we introduce a physical consistency loss based on the wind velocity components. Divergence and vorticity are defined as follows:

$$\nabla \cdot V = \frac{\partial U}{\partial x} + \frac{\partial V}{\partial y}, \quad \omega = \frac{\partial V}{\partial x} - \frac{\partial U}{\partial y}$$

The difference between the inference and true fields is measured using a terrain-weighted physical error metric:

$$L_{\mathrm{phys}} = \left\| M_{\mathrm{terrain}} \odot \left( \nabla \cdot V_{\mathrm{pred}} - \nabla \cdot V_{\mathrm{true}} \right) \right\|_1 + \left\| M_{\mathrm{terrain}} \odot \left( \omega_{\mathrm{pred}} - \omega_{\mathrm{true}} \right) \right\|_1$$

where $M_{\mathrm{terrain}} \in [0,1]^{B \times 1 \times H \times W}$ is an inverse mask generated from terrain gradients:

$$M_{\mathrm{terrain}} = 1 - \frac{\sqrt{\left( \partial_x X_{terrain}["z"] \right)^2 + \left( \partial_y X_{terrain}["z"] \right)^2}}{\max(\cdot)}$$

This method prioritizes physical constraints by assigning higher weights to flatter areas (lower slopes) and reducing weights in complex terrain (higher slopes), which allows for reasonable local deviations. This balance between terrain complexity and energy conservation improves the physical consistency of model outputs. Both the high-frequency and physical loss terms are imposed on the full wind field without masking, as high-frequency structures, divergence, and vorticity are all defined by spatially continuous neighborhood relationships, and masking known observations would compromise the integrity of local gradient patterns and overall dynamical consistency.

Additionally, to suppress high-frequency noise in early training, we apply a gradually increasing ramp schedule to the weight of $L_{\text{high}}$ :

$$\lambda_t = 0.5 \cdot \min\left(1, \frac{t}{T_{\text{ramp}}}\right)$$

where $t$ denotes the current training epoch, and $T_{\text{ramp}}$ represents the ramp period length, ensuring the model first learns stable overall structures before progressively capturing fine-scale variations. The final composite loss function integrates three objectives: pixel consistency, frequency decomposition, and terrain-weighted physical constraints:

$$L_{\text{total}} = \lambda_t L_1 + (1-\lambda_t) L_{\text{high}} + \eta_t L_{\text{phys}}$$

where $\lambda_t$ is the primary loss scheduling coefficient, and $\eta_t$ is the weight for physical constraints. $\eta_t$ linearly increases from zero to its maximum value $\eta_{\max}$ with epoch progression, enabling gradual convergence from numerical fitting to physical consistency.

# 3 Experimental Analysis

All the experiments are conducted on the same hardware, and the relevant parameters are shown in Table 1.

**Table 1** Configurations used in this study.

| Tools | Ubuntu | GPU | Python | Pytorch | Cuda | CPU |
|---|---|---|---|---|---|---|
| Version | 18.04 | RTX 5090 | 3.12 | 2.8.0 | 12.8 | Xeon(R) 8470Q |

## 3.1 Dataset

In this study, we constructed an experimental dataset covering three representative geographical regions to systematically evaluate the model's reconstruction performance and spatial generalization ability under different climatic backgrounds and terrain conditions.

A. Northeast China (38°N–54°N, 116°E–132°E): This region serves as the primary study area and belongs to a typical temperate monsoon climate zone. In terms of terrain, the Greater Khingan Range lies to the west, the Lesser Khingan Range to the north, and the Changbai Mountains to the east, while the central part consists of vast plains, forming an overall horseshoe-shaped topographic pattern surrounded by mountains on three sides and open in the middle. The complex terrain significantly enhances the spatial heterogeneity of near-surface wind fields, making this region highly suitable for studying the evolution characteristics of wind fields under terrain influence. Accordingly, Northeast China was selected as the main training, validation, and testing region for a comprehensive evaluation of the model's reconstruction capability.

B. Europe (42°N–58°N, 4°W–12°E): The European region is generally located within the influence of the mid-latitude westerlies, where large-scale weather systems are frequently active and wind field evolution exhibits pronounced mid-latitude dynamical characteristics. Although the overall terrain in this region is gentler than that of Northeast China, local areas are still jointly affected by orographic lifting and land–sea distribution differences, resulting in wind field spatial structures distinct from those in monsoon climate regions. Using this region as a test area helps evaluate the model's adaptability under different circulation backgrounds.

C. Southeast Asia (10°S–6°N, 114°E–130°E): Southeast Asia is located near the equator and is characterized by persistently high temperature and humidity, with active convective processes and stronger nonlinear features in wind field evolution. At the same time, this region is dominated by fragmented archipelagos, mountainous terrain, and complex coastlines, with pronounced land–sea thermal contrasts and highly variable local circulation processes. Compared with Northeast China and Europe, Southeast Asia is more distinctive in both climatic conditions and terrain structure, thus providing a more challenging test scenario for evaluating the model's cross-climate transfer capability.

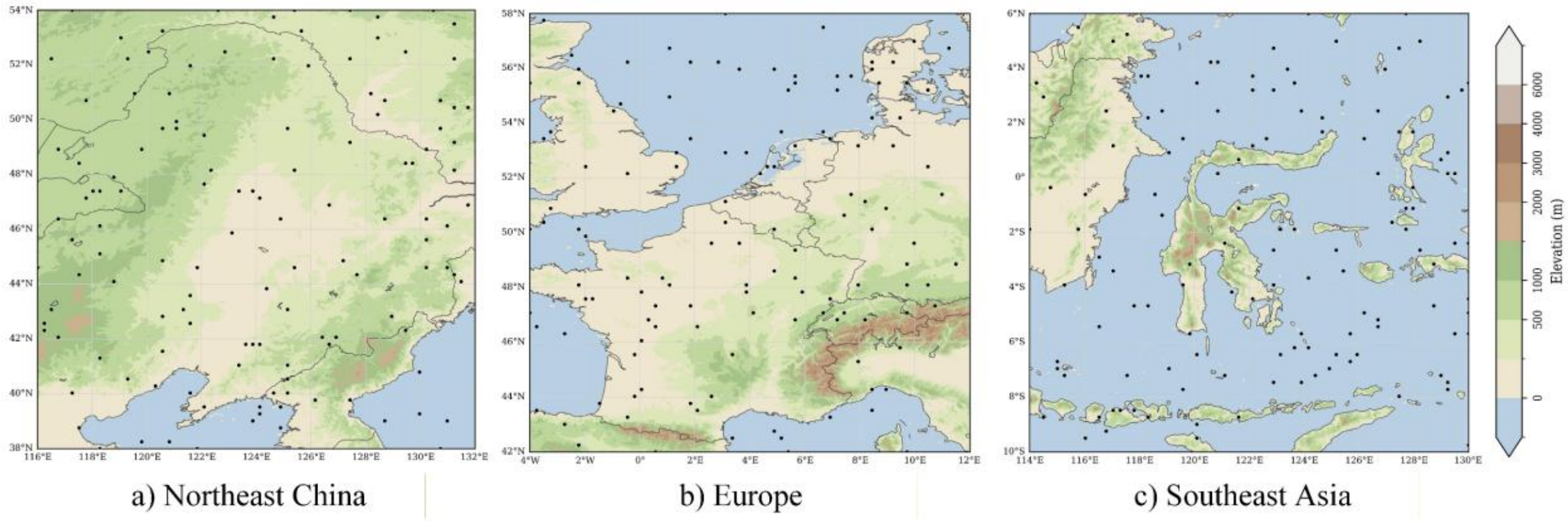

**Figure 3**. Geographical extent and spatial distribution of sparse observation points in the three study areas: (a) Northeast China, (b) Europe, and (c) Southeast Asia. Black dots represent discrete wind speed observation locations, and the background shading represents topographic elevation.

Regarding data construction, we select the u-wind component and v-wind component from the ERA5 reanalysis dataset [50] as the core variables to characterize the wind field. For the temporal split, data from 2020 to 2023 are divided into the training set and validation set at a ratio of 8:2, while data from 2024 are used exclusively as the test set to ensure the temporal independence and objectivity of the test results. To enhance the reproducibility of this study, the complete experimental dataset, including the training, validation, and test sets, has been made publicly available.

During the data preprocessing stage, to reduce the influence of differences in variable scales and improve the stability of model training, all input are standardized using Z-score normalization. The required mean and standard deviation are computed solely from the training set and are then uniformly applied to the validation and test sets, thereby strictly avoiding data leakage. To simulate the sparse, discrete, and irregularly distributed input characteristics of meteorological stations in real observation scenarios on regular-grid data, a random sampling strategy is adopted. Specifically, a limited number of grid points are sampled from the ERA5 grid over the target region as virtual observation stations, and the wind vector information at these locations is used as the model input. In the experiments, three observation densities, 128, 256, and 512 observation points, are considered to evaluate the reconstruction performance of the model under different observation densities. The black dots in Figure. 3 show examples of randomly selected discrete observation points.

## 3.2 Evaluation metrics

To comprehensively evaluate the performance of Zhinv in wind field reconstruction, we adopt the root mean square error (RMSE), mean absolute error (MAE), mean absolute percentage error (MAPE), coefficient of determination ($R^2$), structural similarity index measure (SSIM), weighted circular mean absolute error (wCMAE), and angular accuracy metric ($\mathrm{ACC}_\theta$) for a comprehensive assessment. The definitions of RMSE, MAE, MAPE, and $R^2$ are as follows:

$$\mathrm{RMSE} = \sqrt{\frac{1}{N}\sum_{i=1}^{N}(y_i - \hat{y}_i)^2}$$

$$\mathrm{MAE} = \frac{1}{N}\sum_{i=1}^{N}\left|y_i - \hat{y}_i\right|$$

$$\mathrm{MAPE} = \frac{1}{N'}\sum_{i=1}^{N'}\left|\frac{y_i - \hat{y}_i}{y_i}\right| \times 100\%, \quad y_i > 3$$

$$R^2 = 1 - \frac{\sum_{i=1}^{N}(y_i - \hat{y}_i)^2}{\sum_{i=1}^{N}(y_i - \overline{y})^2}$$

where $y_i$ and $\hat{y}_i$ denote the true and reconstructed wind speeds of the $i$-th sample, respectively, and $N$ is the total number of samples. By restricting the calculation to cases where the true wind speed is greater than 3 m/s, distortion of the metric caused by an excessively small denominator under low-wind-speed conditions can be effectively avoided, thereby making the evaluation results more practically meaningful.

SSIM is used to measure the structural similarity between the reconstructed wind field and the reference wind field, and can comprehensively reflect their consistency in spatial structure:

$$SSIM(y, \hat{y}) = \frac{(2\mu_y\mu_{\hat{y}} + C_1)(2\sigma_{y\hat{y}} + C_2)}{(\mu_y^2 + \mu_{\hat{y}}^2 + C_1)(\sigma_y^2 + \sigma_{\hat{y}}^2 + C_2)}$$

where $\mu$ denotes the mean values of the wind field; $\sigma$ denotes variance; $\sigma_{y\hat{y}}$ denotes covariance; and $C$ is constant introduced to prevent division by zero. The value of SSIM ranges from 0 to 1, with values closer to 1 indicating greater structural similarity between the reconstructed wind field and the reference wind field.

wCMAE is used to evaluate the average error of the model in overall wind direction prediction:

$$\mathrm{wCMAE} = \frac{\sum_{i=1}^{N} w_i |\Delta\theta_i|}{\sum_{i=1}^{N} w_i}$$

where $\Delta\theta_i$ denotes the minimum angular difference between the reconstructed wind direction and the reference wind direction for the $i$-th sample, and $w_i$ is the weighting coefficient, for which the reference wind speed is used as the weight to emphasize the importance of strong-wind samples.

$\mathrm{ACC}_\theta$ is further used to evaluate the model's wind direction inference performance within a small angular error range:

$$\mathrm{ACC}_\theta = \frac{\sum_{i=1}^{N} w_i \mathbb{I}(|\Delta\theta_i| \le \theta)}{\sum_{i=1}^{N} w_i}$$

where $\mathbb{I}$ is the indicator function, which takes the value 1 when the condition is satisfied and 0 otherwise; the weight $w_i$ is the same as that used in wCMAE and is set to the reference wind speed. We choose $\theta = 11.25°$ and $22.5°$ as the thresholds, corresponding to the accuracy within 16-direction and 8-direction wind direction error ranges.

## 3.3 Overall performance analysis

Table 2 presents the quantitative results of different methods for hub-height wind field reconstruction under different numbers of observation points. Overall, as the number of observation points increased from 128 to 512, the performance of all methods improved to varying degrees. This indicates that denser discrete observations can provide more sufficient constraints for regional wind field recovery, thereby reducing reconstruction uncertainty. However, there are clear differences between the methods in their ability to adapt to observation sparsity, which also reflects their fundamental differences in spatial interpolation, structural recovery, and nonlinear feature modeling.

From the overall results, the traditional interpolation methods, Kriging and IDW, consistently performed significantly worse than the deep learning methods. This suggests that relying solely on distance weighting or stationarity assumptions is insufficient to accurately characterize the nonlinear spatial distribution of wind fields under complex terrain conditions. Their relatively poor SSIM and $R^2$ values further indicate that traditional statistical interpolation struggles to recover fine-grained structural information in complex wind fields. Compared with traditional interpolation methods, deep learning-based methods, UNet and PConv, showed clear advantages, demonstrating that data-driven nonlinear mappings can more effectively recover gridded wind fields from discrete observations. It is worth noting that the SSIM values of UNet were only 0.21, 0.13, and 0.18 under the three observation densities, respectively, which were significantly lower than those of the other methods. This indicates that although UNet could achieve relatively good results in terms of pixel-wise error, it still exhibited clear limitations in spatial structural consistency and local texture recovery. Although PConv could alleviate the problem of missing information caused by sparse inputs, its ability to model complex spatial correlations in wind fields remained limited. Moreover, its reconstruction performance deteriorated rapidly as the number of observation points decreased.

The hybrid methods, Kriging+CNN and KCN, combined prior interpolation results with learning-based networks, enabling them to balance global smooth background reconstruction and local nonlinear correction to some extent. Kriging+CNN showed strong competitiveness under all observation settings. For example, when 512 observation points were used, its SSIM reached 0.95, indicating that introducing a convolutional network to refine structural details based on a traditional interpolated field could significantly improve the recovery quality of spatial textures. This suggests that using statistical interpolation to first provide a global field pattern and then employing a convolutional network to resolve local details is an effective strategy for wind field reconstruction. Nonetheless, Zhinv consistently achieved the best performance under all three observation densities, and its advantage remained stable. Under the sparsest condition with only 128 observation points, the MAE, $R^2$, and SSIM of Zhinv reached 0.49 m/s, 0.93, and 0.85, respectively, all outperforming those of the comparison methods. This indicates that even when observations were extremely limited, Zhinv could still effectively recover the main distribution pattern and spatial structure of the regional wind field. As the number of observation points increased to 256 and 512, the performance of Zhinv further improved. Under the 512-point setting, its MAE decreased to 0.20 m/s and SSIM reached 0.97. Zhinv not only showed clear advantages in error but also produced results that were closer to the target field in terms of spatial structure recovery and fitting to the real wind field. This advantage suggests that, compared with traditional interpolation methods and conventional convolution-based reconstruction methods, the proposed method could establish the irregular mapping relationship between discrete observation points and regular grids more effectively. This may be attributed to three factors: (1) its close integration with terrain, as terrain information is tightly incorporated into feature extraction and evolution in the encoder, interpolation module, and decoder, which is crucial for recovering wind field details; (2) its efficient network architecture, in which the graph network can effectively extract sparse features without introducing additional noise, while the interpolation module and decoder are designed in a manner that respects the overall evolution pattern of the wind field; and (3) the effective constraints imposed by the physical loss, which further promote physics-consistent reconstructed wind fields.

**Table 2.** Performance comparison of different methods for hub-height wind field reconstruction in Northeast China. Best results across the different number of points are in **bold**.

| Number of points | Metrics | Kriging | IDW | UNet | PConv | Kriging+CNN | KCN | ZHINV |
|---|---|---|---|---|---|---|---|---|
| 128 | RMSE(m/s) ↓ | 1.42 | 1.95 | 0.80 | 1.59 | 0.80 | 0.99 | **0.77** |
| | MAE(m/s) ↓ | 0.97 | 1.40 | 0.53 | 0.79 | 0.52 | 0.69 | **0.49** |
| | MAPE(%) ↓ | 16.79 | 24.09 | 9.30 | 14.03 | 9.24 | 12.23 | **8.62** |
| | $R^2$ ↑ | 0.75 | 0.53 | 0.92 | 0.69 | 0.92 | 0.88 | **0.93** |

| | SSIM ↑ | 0.58 | 0.47 | 0.21 | 0.53 | 0.83 | 0.73 | **0.85** |
|---|---|---|---|---|---|---|---|---|
| 256 | RMSE(m/s) ↓ | 1.11 | 1.77 | 0.56 | 0.70 | 0.55 | 0.76 | **0.50** |
| | MAE(m/s) ↓ | 0.76 | 1.29 | 0.37 | 0.40 | 0.36 | 0.53 | **0.32** |
| | MAPE(%) ↓ | 13.40 | 22.39 | 6.56 | 7.03 | 6.39 | 9.42 | **5.67** |
| | $R^2$ ↑ | 0.85 | 0.62 | 0.96 | 0.94 | 0.96 | 0.93 | **0.97** |
| | SSIM ↑ | 0.69 | 0.54 | 0.13 | 0.58 | 0.90 | 0.82 | **0.93** |
| 512 | RMSE(m/s) ↓ | 0.90 | 1.70 | 0.39 | 0.44 | 0.37 | 0.58 | **0.33** |
| | MAE(m/s) ↓ | 0.59 | 1.25 | 0.25 | 0.26 | 0.23 | 0.40 | **0.20** |
| | MAPE(%) ↓ | 10.42 | 21.63 | 4.43 | 4.61 | 4.13 | 7.17 | **3.52** |
| | $R^2$ ↑ | 0.90 | 0.65 | 0.98 | 0.98 | 0.98 | 0.96 | **0.99** |
| | SSIM ↑ | 0.80 | 0.59 | 0.18 | 0.70 | 0.95 | 0.89 | **0.97** |

In addition to wind speed magnitude error, wind direction reconstruction accuracy is also an important metric for evaluating the quality of hub-height wind field reconstruction, because wind direction deviations directly affect wind energy assessment and turbine yaw control. To this end, Table 3 further reports the wind direction reconstruction results of different methods under different numbers of observation points. Overall, as the number of observation points increased from 128 to 512, the wCMAE of all methods continuously decreased, indicating that denser discrete observations can effectively reduce the uncertainty of wind direction reconstruction. The traditional interpolation methods, Kriging and IDW, showed relatively weak overall performance, whereas deep learning methods generally exhibited stronger wind direction reconstruction capability. This suggests that data-driven feature extraction, as well as the combination of interpolation priors and convolutional refinement, can effectively improve the reconstruction accuracy of wind direction fields. Under the condition of 128 observation points, the wCMAE of Zhinv was 5.06°, while $ACC_{11.25°}$ and $ACC_{22.5°}$ reached 0.90 and 0.97, respectively, all of which were the best among the compared methods. Under the condition of 512 observation points, the wCMAE of Zhinv further decreased to 1.76°, indicating a high degree of consistency between the predicted and reference wind directions. Overall, Zhinv exhibited lower angular errors and higher directional accuracy under different observation densities, demonstrating that it can more accurately recover the streamline structures and local directional fluctuations of regional wind fields.

**Table 3**. Comparison of wind direction reconstruction performance of different methods in Northeast China. Best results across the different number of points are in **bold**.

| Number of points | Metrics | Kridging | IDW | UNet | PConv | Kriging+CNN | KCN | ZHINV |
|---|---|---|---|---|---|---|---|---|
| 128 | wCMAE(°) | 10.12 | 13.09 | 5.44 | 10.56 | 5.40 | 7.19 | **5.06** |
| | Acc11.25°(%) | 71.89 | 62.95 | 88.63 | 83.09 | 88.74 | 82.72 | **89.84** |
| | Acc22.5°(%) | 89.74 | 84.14 | 96.75 | 90.43 | 96.79 | 94.91 | **97.04** |
| 256 | wCMAE(°) | 7.56 | 11.14 | 3.55 | 4.01 | 3.43 | 5.13 | **3.03** |
| | Acc11.25°(%) | 80.01 | 67.57 | 94.60 | 94.13 | 94.80 | 89.85 | **95.78** |
| | Acc22.5°(%) | 94.06 | 87.88 | 98.84 | 98.26 | 98.90 | 97.69 | **99.10** |
| 512 | wCMAE(°) | 5.62 | 9.93 | 2.25 | 2.35 | 2.10 | 3.81 | **1.76** |
| | Acc11.25°(%) | 87.07 | 70.89 | 97.96 | 97.85 | 98.15 | 94.45 | **98.63** |
| | Acc22.5°(%) | 96.72 | 90.49 | 99.63 | 99.51 | 99.68 | 99.00 | **99.76** |

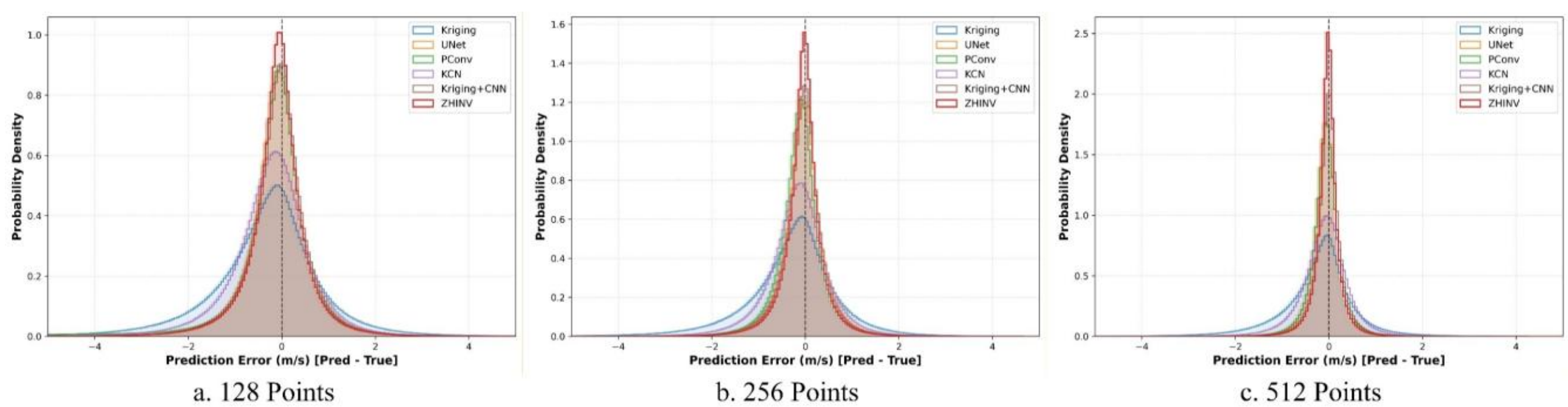

**Figure 4.** Comparison of reconstruction error distributions of different methods under different numbers of observation points (128, 256, and 512). The error distribution of Zhinv is consistently more concentrated around zero and exhibits shorter tails under all three observation densities, indicating lower reconstruction errors.

Figure 4 shows the error distributions of different methods under 128, 256, and 512 observation points. Overall, as the number of observation points increases, the error distributions of all methods gradually converge toward zero, with higher peaks and narrower spreads. This indicates that denser observations can effectively reduce reconstruction errors. The distribution from Kriging exhibits the widest tails, suggesting the largest error dispersion and a higher tendency to produce large deviations. The error distributions of UNet, PConv, and KCN are noticeably more concentrated than that of Kriging, indicating that deep learning methods can reduce overall errors effectively. Kriging+CNN shows higher central peaks and narrower distribution ranges under all three observation densities, demonstrating that the combination of interpolation priors and convolutional correction helps improve error. In contrast, Zhinv exhibits the most peaked and most concentrated distribution, together with the shortest tails on both sides, across the three observation densities. This indicates that its prediction errors are more concentrated around zero, reflecting not only higher overall reconstruction accuracy but also a stronger ability to suppress large-error samples. These results are consistent with the conclusions drawn from the quantitative metrics and spatial error distributions, further confirming that Zhinv consistently achieves superior accuracy under different observation sparsity.

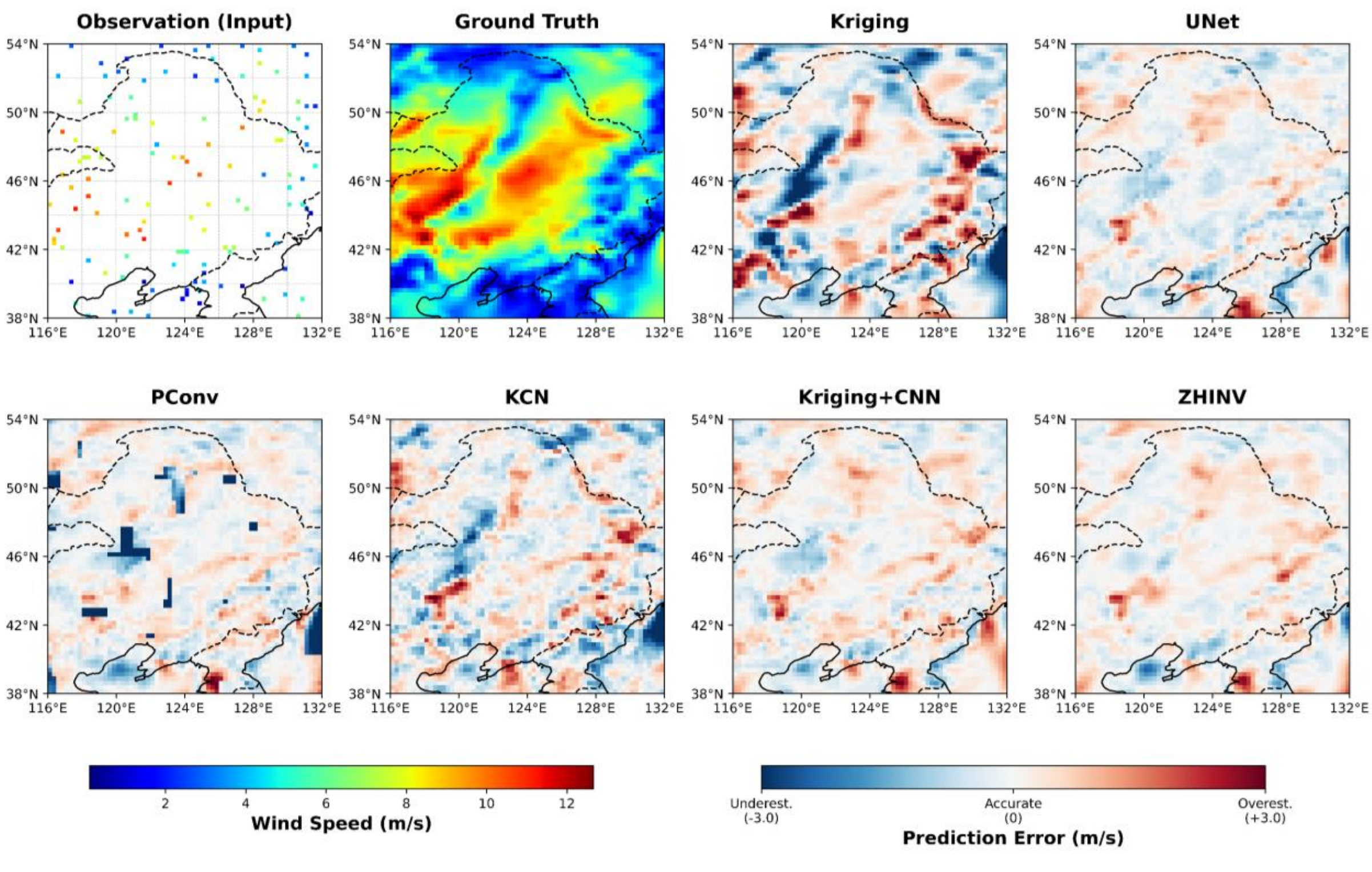

a) 128 Points

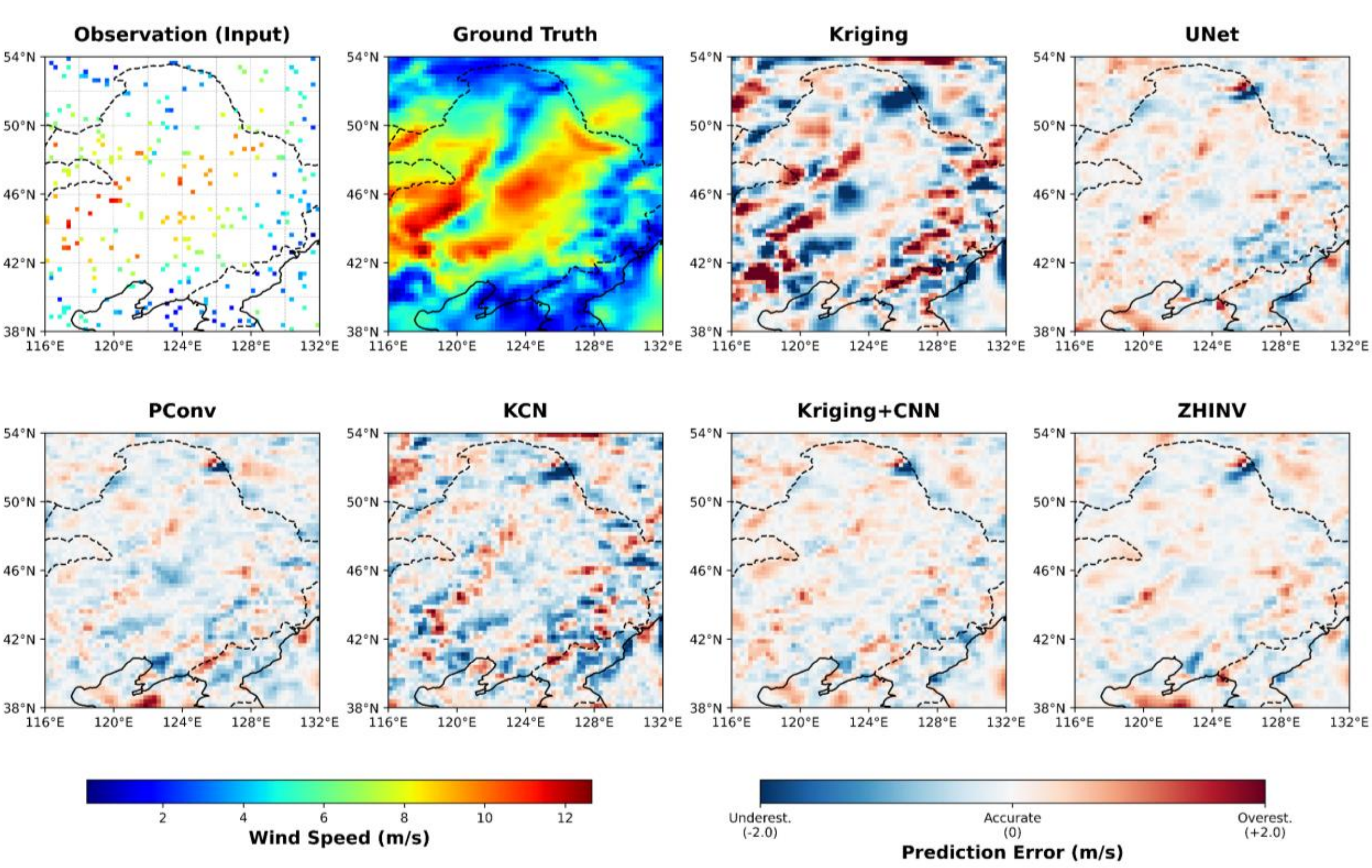

b) 256 Points

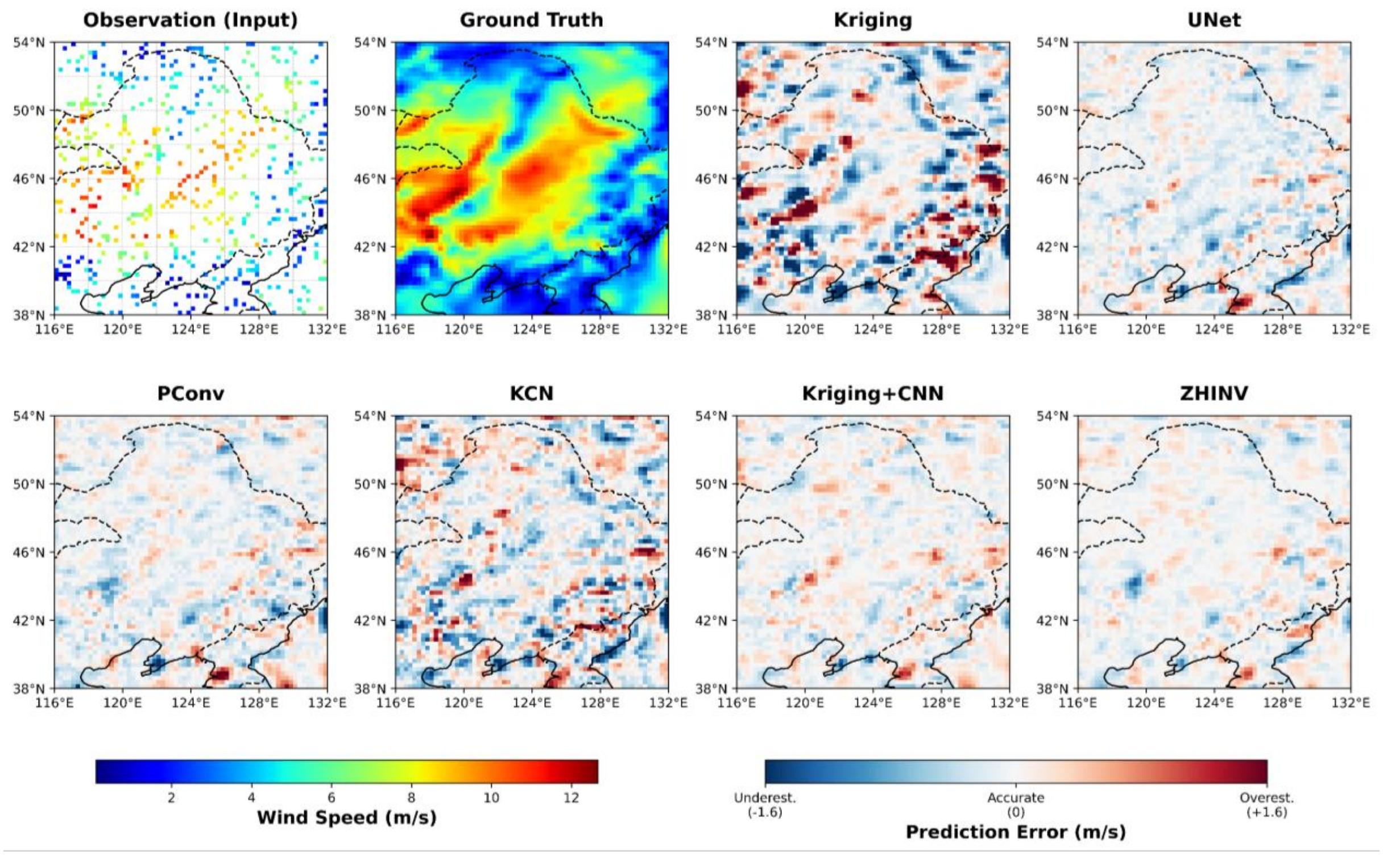

c) 512 Points

**Figure 5.** Comparison of spatial reconstruction and error distributions for various methods under different numbers of observation points. Shown in order are the sparse observation input, the true wind field, and the error distributions of the comparison methods and Zhinv. Zhinv exhibits a smaller and more continuous residual distribution across different observation densities, indicating superior wind field reconstruction accuracy and spatial structure recovery capability.

Figure 5 shows the spatial reconstruction error distributions of different methods for hub-height wind fields under three observation point settings. As the number of observation points increases, denser discrete observations provide stronger spatial constraints and improve the accuracy of wind field reconstruction. The error magnitudes of all methods decrease, and the error distributions gradually change from large-scale and highly scattered patterns to more localized and weaker disturbances. Although Kriging can recover large-scale smooth structures, it remains unable to accurately capture local variations in complex wind fields. As a result, it consistently exhibits large areas of positive and negative biases, especially in regions with strong wind speed gradients. The error distributions of UNet and PConv are reduced, but noticeable local residuals remain. PConv shows pronounced block-like artifacts under the 128-point condition, reflecting its limited ability to recover structures under extremely sparse observations. The error distribution of Kriging+CNN is significantly improved, indicating that the combination of interpolation priors and convolutional correction can effectively improve the overall reconstruction quality. In contrast, Zhinv exhibits the smallest and most uniform error distribution under all three observation densities. Especially in the high-wind-speed core regions and their surrounding transition zones, its residual errors are weaker and its spatial continuity is better, demonstrating that it can more stably recover both the dominant wind field structure and local details under different sparse observation conditions.

## 3.4 Ablation experiment

**Table 4.** Ablation results under different numbers of sparse observation points in Northeast China. Best results across the different number of points are in **bold**.

| Number of points | Model | RMSE (m/s) | MAE (m/s) | MAPE | $R^2$ | SSIM | wCMAE(°) |
|---|---|---|---|---|---|---|---|
| 128 | ZHINV-Ele | 0.80 | 0.53 | 9.45 | 0.92 | 0.83 | 5.41 |
| | ZHINV-Loss | 0.74 | 0.48 | 8.53 | 0.93 | **0.86** | 4.85 |
| | ZHINV | **0.73** | **0.47** | **8.32** | **0.94** | **0.86** | **4.74** |
| 256 | ZHINV-Ele | 0.57 | 0.37 | 6.63 | 0.96 | 0.90 | 3.57 |
| | ZHINV-Loss | 0.50 | 0.32 | 5.77 | **0.97** | 0.92 | 3.06 |
| | ZHINV | **0.49** | **0.31** | **5.50** | **0.97** | **0.93** | **2.94** |
| 512 | ZHINV-Ele | 0.37 | 0.24 | 4.24 | 0.98 | 0.96 | 2.16 |
| | ZHINV-Loss | **0.31** | 0.20 | 3.51 | **0.99** | **0.97** | 1.74 |
| | ZHINV | **0.31** | **0.19** | **3.43** | **0.99** | **0.97** | **1.68** |

Table 4 presents the ablation results under different observation point settings. Compared with the variants without terrain information or the specialized loss (MFPH-Loss), the full Zhinv outperforms its variants across all metrics, indicating that both terrain information and the customized loss function play important roles in wind field reconstruction performance. After removing terrain information, a significant performance decay is observed in all metrics. Under the 512-observation-point setting, the RMSE degrades by as much as 20%, demonstrating that terrain priors are crucial for characterizing local wind speed differences and flow structure. After introducing the improved loss function, Zhinv shows more pronounced improvements in SSIM and wind direction-related metrics. This indicates that the loss function not only reduces error magnitude, but also enhances spatial structural fidelity and directional consistency. These ablation results further verify the synergistic contribution of terrain information and loss function design in improving reconstruction accuracy and stability.

## 3.5 Transfer performance analysis

**Table 5.** Cross-regional transfer results under the constant-density setting. The model was trained on Northeast China and then transferred to Europe and Southeast Asia for testing after one epoch of fine-tuning. Best results are in **bold**.

| Regions | Number of points | Metrics | Kriging | UNet | KCN | ZHINV |
|---|---|---|---|---|---|---|
| Europe | 128 | RMSE ↓ | 1.52 | 1.06 | 1.48 | **0.92** |
| | | $R^2$ ↑ | 0.87 | 0.94 | 0.87 | **0.95** |
| | | SSIM ↑ | 0.64 | 0.19 | 0.65 | **0.84** |
| | | wCMAE ↓ | 8.27 | 5.56 | 8.09 | **4.95** |
| | 256 | RMSE ↓ | 1.20 | 0.75 | 1.07 | **0.61** |
| | | $R^2$ ↑ | 0.92 | 0.97 | 0.93 | **0.98** |

| | | SSIM ↑ | 0.74 | 0.11 | 0.75 | **0.91** |
|---|---|---|---|---|---|---|
| | | wCMAE ↓ | 5.94 | 3.92 | 5.68 | **3.08** |
| | 512 | RMSE ↓ | 0.93 | 0.55 | 0.83 | **0.41** |
| | | $R^2$ ↑ | 0.95 | 0.98 | 0.96 | **0.99** |
| | | SSIM ↑ | 0.84 | 0.14 | 0.84 | **0.96** |
| | | wCMAE ↓ | 4.14 | 2.57 | 4.15 | **1.86** |
| Southeast Asia | 128 | RMSE ↓ | 1.39 | 0.80 | 1.17 | **0.73** |
| | | $R^2$ ↑ | 0.72 | 0.91 | 0.80 | **0.92** |
| | | SSIM ↑ | 0.47 | 0.21 | 0.57 | **0.80** |
| | | wCMAE ↓ | 9.65 | 5.93 | 8.31 | **5.16** |
| | 256 | RMSE ↓ | 1.19 | 0.62 | 0.87 | **0.52** |
| | | $R^2$ ↑ | 0.79 | 0.94 | 0.89 | **0.96** |
| | | SSIM ↑ | 0.58 | 0.12 | 0.72 | **0.88** |
| | | wCMAE ↓ | 7.72 | 4.24 | 5.96 | **3.49** |
| | 512 | RMSE ↓ | 0.98 | 0.48 | 0.76 | **0.39** |
| | | $R^2$ ↑ | 0.86 | 0.97 | 0.91 | **0.98** |
| | | SSIM ↑ | 0.71 | 0.21 | 0.80 | **0.93** |
| | | wCMAE ↓ | 6.06 | 3.10 | 4.67 | **2.37** |

**Table 6.** Comparison of cross-regional transfer performance across different regions and different numbers of sparse observation points. The model pretrained in Northeast China with 512 observation points was transferred to Europe and Southeast Asia under the 128- and 256-point settings, respectively, with one epoch of fine-tuning. Best results are in **bold**.

| Regions | Number of points | Metrics | Kriging | UNET | KCN | ZHINV |
|---|---|---|---|---|---|---|
| Europe | 256 | RMSE ↓ | 1.20 | 0.76 | 1.18 | **0.63** |
| | | $R^2$ ↑ | 0.92 | 0.97 | 0.92 | **0.98** |
| | | SSIM ↑ | 0.74 | 0.11 | 0.72 | **0.90** |
| | | wCMAE ↓ | 5.94 | 3.88 | 6.40 | **3.18** |
| | 128 | RMSE ↓ | 1.52 | 1.05 | 1.64 | **0.94** |
| | | $R^2$ ↑ | 0.87 | 0.94 | 0.85 | **0.95** |
| | | SSIM ↑ | 0.64 | 0.18 | 0.59 | **0.83** |
| | | wCMAE ↓ | 8.27 | 5.98 | 9.44 | **5.15** |
| Southeast Asia | 256 | RMSE ↓ | 1.19 | 0.62 | 0.93 | **0.55** |
| | | $R^2$ ↑ | 0.79 | 0.94 | 0.87 | **0.96** |
| | | SSIM ↑ | 0.58 | 0.12 | 0.71 | **0.88** |
| | | wCMAE ↓ | 7.72 | 4.24 | 6.26 | **3.62** |
| | 128 | RMSE ↓ | 1.39 | 0.84 | 1.21 | **0.74** |
| | | $R^2$ ↑ | 0.72 | 0.90 | 0.79 | **0.92** |
| | | SSIM ↑ | 0.47 | 0.19 | 0.56 | **0.79** |

| | | wCMAE ↓ | 9.65 | 5.86 | 8.72 | **5.31** |
|---|---|---|---|---|---|---|

To further verify the generalization capability of the model, we conducted cross-regional transfer experiments. Specifically, the model trained in Northeast China was fine-tuned for only one epoch and then used for inference in Europe and Southeast Asia, respectively. Considering that, in practical applications, not only the regional distribution but also the number of observation points may vary, we designed two types of experiments, namely constant-density transfer and cross-density transfer, to simultaneously evaluate the model's adaptability to changes in both region and observation sparsity.

As shown in Table 5, consistent with the results in Northeast China, denser observation inputs effectively reduced the uncertainty in cross-regional inference. In both Europe and Southeast Asia, as the number of observation points increased, the RMSE and wCMAE of all methods continuously decreased, while $R^2$and SSIM continuously improved. Under the constant-density transfer setting, Zhinv achieved the best results in both regions across all three density of observations. For example, under the 512 setting in Europe, the RMSE, SSIM, and wCMAE of Zhinv reached 0.41 m/s, 0.96, and 1.86°, respectively. Under the corresponding setting in Southeast Asia, these metrics were 0.39 m/s, 0.93, and 2.37°, respectively, all of which were clearly better than those of the other methods. Even under the 128-point setting, Zhinv still maintained the lowest errors and the highest structural similarity, indicating that it could still adapt well to the wind field distribution characteristics of different regions given extremely sparse observations.

Different-point-number transfer further increased the difficulty of the task, because the model needed to maintain stable inference performance when the input observation density changed. The results in Table 6 show that when the model trained under the 512-point setting was used to infer the 256-point and 128-point scenarios, the accuracy of all methods declined to some extent, but Zhinv still remained superior under all settings. In Europe, when the 512-point Northeast China model was used for inference under the 256-point and 128-point settings, the RMSE values of Zhinv were 0.63 m/s and 0.94 m/s, respectively, both lower than those of the comparison methods, while its $R^2$and SSIM still remained the highest. In Southeast Asia, Zhinv also achieved lower RMSE and wCMAE values under the two settings, showing more stable cross-point-number adaptability. These results indicate that Zhinv does not simply rely on fitting to a specific region or a fixed observation density, but can learn a robust and generalized mapping relationship from discrete observations to fine-grid wind fields. As a result, it can still maintain high reconstruction accuracy, structural fidelity, and directional consistency under cross-regional and cross-point-number conditions.

## 3.6 Resilience under extreme meteorological boundaries

**Table 7.** Comparison of the MAE of different methods under extreme meteorological conditions with 512 observation points. Here, "Overall" denotes the MAE over all samples; "P95Frame" and "P99Frame" denote the 95th and 99th percentiles of the frame-wise error distribution, respectively; "Worst1%Frame" denotes the average MAE over the worst 1% of frames; Complex-Flow denotes the MAE in regions with complex flow patterns; and High-Speed denotes the MAE under high-wind-speed conditions. Best results are in **bold**.

| Model | Overall | P95 Frame | P99 Frame | Worst 1% Frame | Complex-Flow | High-Speed |
|---|---|---|---|---|---|---|
| Kriging | 0.60 | 0.93 | 1.12 | 1.24 | 0.61 | 0.61 |
| CNN | 0.25 | 0.35 | 0.40 | 0.43 | 0.28 | 0.27 |
| PConv | 0.25 | 0.35 | 0.41 | 0.43 | 0.28 | 0.27 |
| KCN | 0.40 | 0.51 | 0.59 | 0.61 | 0.45 | 0.44 |
| Kriging+CNN | 0.23 | 0.33 | 0.38 | 0.41 | 0.24 | 0.24 |
| ZHINV | **0.19** | **0.28** | **0.32** | **0.34** | **0.21** | **0.21** |

Wind speed reconstruction requires both high overall accuracy, and robustness to more extreme meteorological conditions, complex flow fields, and key turbine operating wind speed intervals. Hence, we further evaluated the methods from three aspects: long-tail error, robustness under complex scenarios, and accuracy within key wind speed intervals. Here, we selected the reconstruction results of Zhinv and the other models under the 512-observation-point setting for evaluation. P95 Frame MAE, P99 Frame MAE, and Worst 1% Frame MAE are used to characterize the reliability of the models on long-tail and extreme samples. Complex-Flow MAE and High-Speed MAE represent the average errors on complex-flow samples and high-wind-speed samples, respectively, and are used to assess the reconstruction capability of the models under challenging scenarios.

The results in Table 7 show that, under the 512-observation-point setting, Zhinv achieved the best performance on all frame-level metrics. Its Worst 1% Frame MAE, Complex-Flow MAE, and High-Speed MAE were 0.34 m/s, 0.21 m/s, and 0.21 m/s, respectively. This indicates that Zhinv not only yields the lowest overall error, but also exhibits stronger robustness on long-tail samples, complex flow fields, and high-wind-speed conditions. In contrast, Kriging performed the worst on all metrics. Although CNN, PConv, and KCN showed clear improvements over traditional interpolation methods, their performance degradation on difficult samples remained evident. Kriging+CNN was the strongest baseline, but it was still consistently outperformed by Zhinv across all core metrics.

To further analyze model performance within key turbine operating wind speed intervals, we divided the wind speeds at unobserved grid points into five intervals according to the 5 MW turbine power curve and the reference wind speed. As shown in Table 8, Zhinv achieved the lowest MAE among all methods in all five intervals. This demonstrates that the advantage of Zhinv is not limited to a reduction in overall average error, but also extends to maintaining higher reconstruction accuracy near the cut-in wind speed, near the rated wind speed, and in higher wind speed intervals. Overall, under the 512-observation-point setting, Zhinv not only achieved the lowest overall error, but also showed stronger reliability and applicability in long-tail scenarios, complex scenarios, and key wind speed intervals.

**Table 8.** Comparison of MAE under different turbine-relevant operating wind speed intervals. According to the reference wind speed at unobserved grid points, the samples are divided into five intervals: Near cut-in (3–4 m/s), Sub-rated (4–10.4 m/s), Near rated (10.4–12.4 m/s), Above rated (12.4–23 m/s), and Near cut-out (23–25

m/s). The interval division is based on the key operating stages of the adopted generic 5 MW turbine power curve, including the cut-in, rated-transition, above-rated, and near cut-out regions. Best results are in **bold**.

| Model | Near cut-in (m/s) | Sub-rated (m/s) | Near rated (m/s) | Above rated (m/s) | Near cut-out (m/s) |
|---|---|---|---|---|---|
| Kriging | 0.55 | 0.60 | 0.87 | 1.11 | 1.21 |
| CNN | 0.25 | 0.24 | 0.31 | 0.37 | 0.54 |
| PConv | 0.26 | 0.24 | 0.30 | 0.36 | 0.52 |
| KCN | 0.39 | 0.39 | 0.51 | 0.64 | 1.35 |
| Kriging+CNN | 0.24 | 0.22 | 0.25 | 0.29 | 0.45 |
| ZHINV | **0.20** | **0.19** | **0.22** | **0.27** | **0.40** |

## 3.7 Robustness to observation-node failures

To evaluate model robustness under observation-node failures, we progressively removed a subset of sparse input nodes during inference. The experiments were conducted with 512, 256, and 128 initial observation nodes, and the maximum removal ratio was set to 25% in each case, corresponding to 128, 64, and 32 removed nodes. As shown in Figure 6, the MAE of both Zhinv and Kriging+CNN increases as more nodes are removed, indicating that observation-node failures weaken the spatial constraints required for regional wind field reconstruction. Under the same missing-node conditions, Zhinv maintains lower absolute errors in most cases, particularly under the 256- and 128-node settings, suggesting that it can still provide practically useful reconstruction accuracy after partial observation loss. Notably, a lower absolute MAE does not necessarily imply a smaller relative degradation from the no-failure baseline. Kriging+CNN shows stronger conservative stability in some high-missing-ratio cases. For example, when 64 and 128 nodes are removed under the 512-node setting, its MAE is slightly lower than that of Zhinv. This may be attributed to the spatial smoothing prior introduced by Kriging interpolation, which can help reduce the overall average error when observational constraints become insufficient. In contrast, Zhinv relies more on structural correlations among the remaining observation nodes and terrain-guided reconstruction for fine-grained field recovery. It is therefore more likely to maintain lower errors under mild to moderate node failures, but can also be affected when the available observational information becomes severely limited. The horizontal dashed line in Figure 6 shows that even after some observation nodes failed, Zhinv still had a significantly lower error than pure Kriging across all test settings. This indicates that Zhinv's advantage does not only come from the input conditions when the observation points are relatively complete, but also from its ability to recover a more accurate wind field than traditional spatial interpolation methods even when observations are incomplete. Overall, Fig. 6 shows that Zhinv achieves lower reconstruction errors in most node-failure scenarios, while Kriging+CNN retains a degree of smoothing-based stability under severe observation loss, reflecting different degradation mechanisms of the two methods.

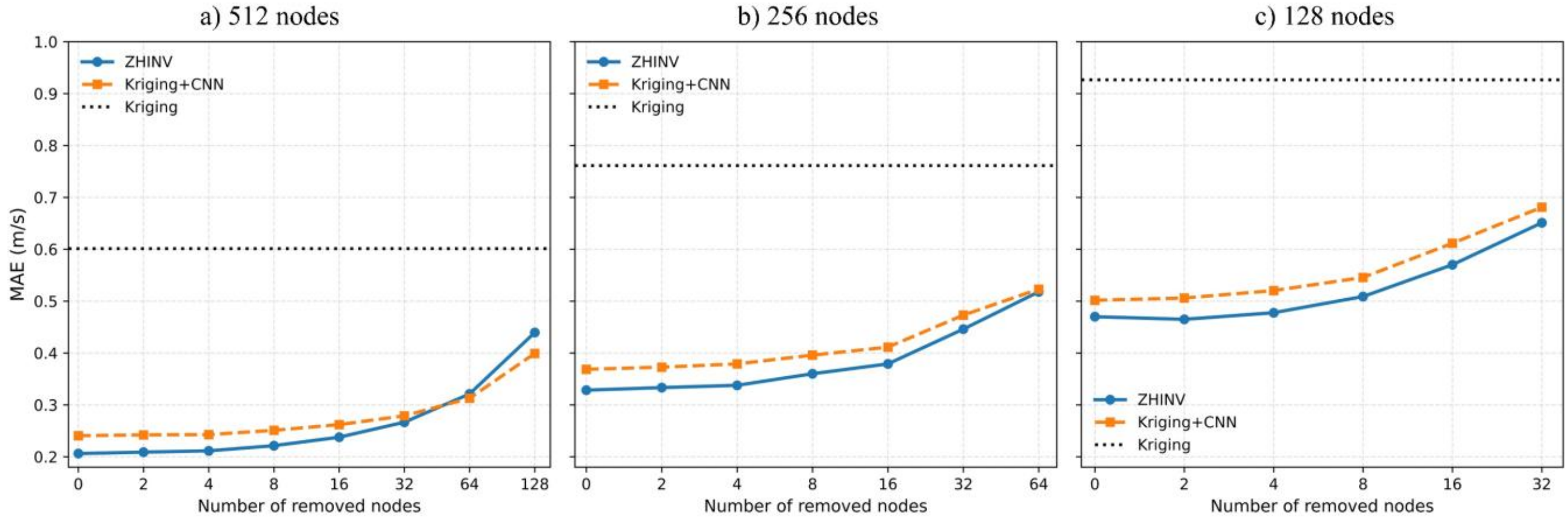

**Figure 6.** Robustness comparison under observation-node failures. MAE of Zhinv and Kriging+CNN under different numbers of removed observation nodes for the 512-, 256-, and 128-node settings. The maximum removal ratio is 25%. Horizontal dashed lines indicate the accuracy of pure Kriging under complete observation nodes.

## 3.8 Inference efficiency test

**Table 9.** Model parameters, computational cost, and inference time under different numbers of sparse observation points.

| Number of points | Parameters (M) | GFlops (G) | Time (ms) |
|---|---|---|---|
| 128 | 6.22 | 49 | 8 |
| 256 | 6.22 | 51 | 12 |
| 512 | 6.22 | 55 | 21 |

The efficiency results in Table 9 show that Zhinv maintains architectural consistency and real-time inference capability under different observation-point configurations. As the number of input observation points increases from 128 to 256 and 512, the number of model parameters remains constant at 6.22 M, indicating that the model size does not expand with the increase in input points. The additional observations only increase the forward computational cost, without significantly increasing the model storage burden. Meanwhile, the computational complexity increases only slightly from 49 to 51 and 55 GFLOPs, respectively, suggesting that the proposed method has good complexity control with respect to changes in observation density. From the perspective of practical deployment, the single-frame inference time is 8 ms, 12 ms, and 21 ms, respectively. Although the inference latency increases as the number of input points grows, millisecond-level inference is still achieved under the 512-point setting, demonstrating good real-time performance. Overall, Zhinv achieves a favorable balance among model size, computational complexity, and inference efficiency while maintaining strong adaptability to different input densities, indicating its practical potential for rapid regional wind field reconstruction.

# 4 Discussion

This study focuses on the problem of rapidly constructing gridded regional wind fields from sparse discrete observations and proposes Zhinv, an efficient regional wind field reconstruction solution for tasks such as wind power regulation, wind resource assessment, and low-altitude environmental sensing. Unlike numerical weather prediction and CFD methods that rely on solving complete dynamical processes, Zhinv is designed for a task that is closer to real-time applications, namely directly recovering regular-grid regional wind fields from limited observations at the current time. Zhinv integrates graph-structured sparse encoding, terrain-constrained modulation, and regular-grid reconstruction into a unified framework, enabling the model to simultaneously capture the spatial correlations among discrete observations, the influence of terrain heterogeneity, and the demand for continuous representation on the target grid. Experimental results show that Zhinv not only outperforms traditional interpolation methods and comparative deep learning models in terms of overall error, but also exhibits strong accuracy in more challenging scenarios, including complex-flow regions, high-wind-speed intervals, and long-tail error cases. Transfer experiments across different regions and different observation densities further demonstrate the model's superior generalization capability. In addition, the model achieves high inference efficiency, requiring only about 20 ms to complete wind field reconstruction over Northeast China on an RTX 5090, which lays a solid foundation for its practical rapid deployment.

Nevertheless, this study still has several limitations and sources of error. First, the sparse observations in this work are mainly constructed through random sampling from the reference wind field. Although this setting is helpful for systematically analyzing model performance under different point densities, it still does not fully account for complex factors in real operational observation networks, such as uneven station distribution, terrain-biased deployment, observational noise, and long-term missing data. Second, the reference wind field used for model training and evaluation is itself subject to resolution limitations. Although it can reasonably represent wind speed distributions at the regional scale, it may still be insufficient for capturing finer-scale local disturbances, terrain-induced small-scale flows, and strongly nonstationary structures, which may further limit the model's ability to learn real wind field details. Third, Zhinv currently relies mainly on sparse wind speed observations and terrain information for reconstruction, and does not yet explicitly incorporate more complete atmospheric physical mechanisms, such as vertical wind field structure, boundary layer processes, thermal stratification, humidity variations, and local turbulence. Therefore, under complex weather conditions, extreme wind situations, or in regions far from observation points, reconstruction biases may still occur. In addition, it should be emphasized that Zhinv is essentially a current wind field reconstruction model rather than a future evolution prediction model. It addresses the problem of recovering the current regional wind field from current discrete observations, and is therefore more suitable as a rapid field reconstruction module than as an independent forecasting system.

Future work can be extended in several directions. On the one hand, more systematic validation should be carried out under real observation network conditions, with a focus on evaluating the robustness and generalization capability of the model under non-uniform station layouts, observational noise, missing-data scenarios, and cross-regional transfer conditions, thereby improving its credibility for operational applications. On the other hand, Zhinv can be combined with upstream station or sensor network forecasting models, thereby presenting a pathway from discrete point-level predictions to gridded regional wind field prediction, and extending its role from current wind field reconstruction to future wind field prediction. In addition, future studies may further enhance the model's multi-source information fusion and physical constraint capabilities, for example by incorporating multi-height wind fields, boundary layer information, surface properties, or cross-scale structural constraints, so as to improve reconstruction accuracy and physical consistency under complex conditions. Overall, Zhinv demonstrates that, even without relying on a complete numerical solver (e.g. CFD or NWP), it is possible to achieve rapid and spatially consistent regional wind field construction by utilizing the coupling between sparse observations, terrain conditions, and spatial positions. This not only provides a feasible solution for real-time intelligent wind field sensing, but also offers new insight into future meteorological intelligent modeling research aimed at moving from point-level observations and point-level predictions toward regional field-level representations.

# 5 Conclusion

This study proposes the Zhinv model to address the need for rapidly constructing gridded regional wind fields from sparse discrete observations. By integrating graph-structured sparse encoding, terrain constraints, and regular-grid reconstruction, the proposed method achieves end-to-end reconstruction from irregular discrete observations to gridded hub-height wind fields. Experimental results show that Zhinv performs well under different observation-point settings and achieves strong results in RMSE, MAE, MAPE, $R^2$, SSIM, and wind direction reconstruction metrics. Moreover, Zhinv remains stable and robust in complex-flow regions, high-wind-speed samples, long-tail error samples, and key turbine operating wind speed intervals, consistently reducing overall error and effectively recovering the spatial structure and local details of wind fields. Cross-regional experiments further show that Zhinv maintains good reconstruction performance in both Europe and Southeast Asia, demonstrating its generalization capability. Overall, the reconstruction performance of Zhinv demonstrates that, even without relying on a complete numerical solver, it is still possible to achieve rapid and spatially consistent regional wind field construction by learning the relationships among discrete observations, terrain conditions, and spatial positions. The model can provide efficient regional wind field support for tasks such as wind power regulation, wind resource assessment, and low-altitude environmental sensing. In future work, real observation networks, multi-source meteorological information, and stronger physical constraints will be further incorporated to improve the model's accuracy and applicability to complex real-world operational scenarios.

# References:

[1] Zhao, J., Li, F. & Zhang, Q. Impacts of renewable energy resources on the weather vulnerability of power systems. **Nat. Energy 9**, 1407–1414 (2024).

[2] Wang, Y., Wang, R., Tanaka, K. *et al.* Global spatiotemporal optimization of photovoltaic and wind power

to achieve the Paris Agreement targets. **Nat. Commun. 16**, 2127 (2025).

[3] System impacts of wind energy developments: Key research challenges and opportunities. **Joule**.

[4] Regional coordination can alleviate the cost burden of a deeply decarbonized energy system.

[5] Ding, J. W. & Hsieh, I. Y. L. Super-resolution wind mapping with deep learning for scalable renewable energy planning. **Commun. Earth Environ.** (2025).

[6] Zhang, Y., Zhao, Y. *et al.* Three-dimensional spatiotemporal wind field reconstruction based on physics-informed deep learning. **Appl. Energy 300**, 117390 (2021).

[7] Bauer, P., Thorpe, A. & Brunet, G. The quiet revolution of numerical weather prediction. **Nature 525**, 47–55 (2015).

[8] Bauer, P., Magnusson, L., Thépaut, J.-N. & Hamill, T. M. Aspects of ECMWF model performance in polar areas. **Q. J. R. Meteorol. Soc.** (2016).

[9] Fujii, K. Progress and future prospects of CFD in aerospace—wind tunnel and beyond. **Prog. Aerosp. Sci. 41**, 455–470 (2005).

[10] Li, J. & Heap, A. D. Spatial interpolation methods applied in the environmental sciences: A review. **Environ. Model. Softw. 53**, 173–189 (2014).

[11] Houndekindo, F. *et al.* Machine learning and statistical approaches for wind resource assessment at partially sampled and unsampled locations: A review. **Energy Convers. Manag.** (2025).

[12] Liu, Y., Wang, R., Gu, Y. *et al.* Physics-inspired and data-driven two-stage deep learning approach for wind field reconstruction with experimental validation. **Energy 298**, 131230 (2024).

[13] Zhou, Y., Lin, C., Kikumoto, H. *et al.* Rooftop wind field reconstruction using sparse sensors: From deterministic to generative learning methods. **Build. Environ.** 114480 (2026).

[14] Wu, Y., Zhuang, D., Labbé, A. *et al.* Inductive graph neural networks for spatiotemporal kriging. In **Proc. AAAI Conf. Artif. Intell. 35**, 4478–4485 (2021).

[15] Skamarock, W. C., Klemp, J. B., Dudhia, J. *et al.* A description of the advanced research WRF version 4. **NCAR Tech. Note** NCAR/TN-556+STR (2019).

[16] Bi, K. *et al.* Accurate medium-range global weather forecasting with 3D neural networks. **Nature** (2023).

[17] Zhang, Z., Lin, L., Gao, S. *et al.* Wind speed prediction in China with fully-convolutional deep neural network. **Renew. Sustain. Energy Rev. 201**, 114623 (2024).

[18] Sanderse, B., van der Pijl, S. P. & Koren, B. Review of computational fluid dynamics for wind turbine wake aerodynamics. **Wind Energy 14**, 799–819 (2011).

[19] Porté-Agel, F., Bastankhah, M. & Shamsoddin, S. Wind-turbine and wind-farm flows: A review. **Bound.-Layer Meteorol. 174**, 1–59 (2020).

[20] Lin, C., Tie, R., Yi, S. *et al.* Reconstructing fine-scale 3D wind fields with terrain-informed machine learning. **Nat. Commun.** (2026).

[21] Baitureyeva, A., Yang, T. & Wang, H. S. Development of machine learning-aided rapid CFD prediction for optimal urban wind environment design. **Sustain. Cities Soc. 121**, 106208 (2025).

[22] Shepard, D. A two-dimensional interpolation function for irregularly-spaced data. In **Proc. 1968 23rd ACM Natl. Conf.**, 517–524 (1968).

[23] Chen, W., Li, Y., Reich, B. J. & Sun, Y. DeepKriging: Spatially dependent deep neural networks for spatial prediction. **Stat. Sin. 34**, 291–311 (2024).

[24] Luo, W., Taylor, M. C. & Parker, S. R. A comparison of spatial interpolation methods to estimate continuous wind speed surfaces using irregularly distributed data from England and Wales. **Int. J. Climatol. 28**, 947–959 (2008).

[25] Workneh, H. T., Chen, X., Ma, Y. *et al.* Comparison of IDW, Kriging and orographic based linear interpolations of rainfall in six rainfall regimes of Ethiopia. **J. Hydrol. Reg. Stud. 52**, 101696 (2024).

[26] Chen, G. *et al.* Spatial estimation of wind speed: A new integrative model using inverse distance weighting and power law. **Int. J. Digit. Earth 9**, 1184–1200 (2016).

[27] Appleby, G., Liu, L. & Liu, L. P. Kriging convolutional networks. In **Proc. AAAI Conf. Artif. Intell. 34**, 3187–3194 (2020).

[28] Bochow, N., Poltronieri, A., Rypdal, M. *et al.* Reconstructing historical climate fields with deep learning. **Sci. Adv. 11**, eadp0558 (2025).

[29] Schmutz, Y., Imfeld, N., Brönnimann, S. *et al.* Enhanced video inpainting: A deep learning approach for historical weather reconstruction. **J. Geophys. Res. Mach. Learn. Comput. 1**, e2024JH000299 (2024).

[30] Eusebi, R., Vecchi, G. A., Lai, C. Y. *et al.* Realistic tropical cyclone wind and pressure fields can be reconstructed from sparse data using deep learning. **Commun. Earth Environ. 5**, 8 (2024).

[31] Lam, R. *et al.* Learning skillful medium-range global weather forecasting. **Science** (2023).

[32] Pathak, J. *et al.* FourCastNet: A global data-driven high-resolution weather model using adaptive Fourier neural operators. arXiv (2022).

[33] Price, I., Sanchez-Gonzalez, A., Alet, F. *et al.* Probabilistic weather forecasting with machine learning. **Nature 637**, 84–90 (2025).

[34] Chen, L., Zhong, X., Zhang, F. *et al.* FuXi: A cascade machine learning forecasting system for 15-day global weather forecast. **npj Clim. Atmos. Sci. 6**, 190 (2023).

[35] Zhang, Z., Lin, L., Wei, S. *et al.* DUSTFN: A dual-path U spatiotemporal fusion network for wind speed prediction. **Inf. Fusion** 103504 (2025).

[36] Zhong, X. *et al.* FuXi-ENS: A machine learning model for efficient and accurate ensemble weather prediction. **Sci. Adv. 11**, 2854 (2025).

[37] Geng, X., Xu, L., He, X. *et al.* Graph optimization neural network with spatio-temporal correlation learning for multi-node offshore wind speed forecasting. **Renew. Energy 180**, 1014–1025 (2021).

[38] Stańczyk, T. & Mehrkanoon, S. Deep graph convolutional networks for wind speed prediction. In **ESANN 2021 Proc.** 147–152 (2021).

[39] Bentsen, L. Ø., Warakagoda, N. D., Stenbro, R. *et al.* Spatio-temporal wind speed forecasting using graph networks and novel transformer architectures. **Appl. Energy 333**, 120565 (2023).

[40] Gao, Z., Li, Z., Xu, L. *et al.* Dynamic adaptive spatio-temporal graph neural network for multi-node offshore wind speed forecasting. **Appl. Soft Comput. 141**, 110294 (2023).

[41] Yan, B., Shen, R., Li, K. *et al.* Spatio-temporal correlation for simultaneous ultra-short-term wind speed prediction at multiple locations. **Energy 284**, 128418 (2023).

[42] Zhang, Z., Lin, L., Gao, S. *et al.* A machine learning model for hub-height short-term wind speed prediction. **Nat. Commun. 16**, 3195 (2025).

[43] Santos, J. E., Fox, Z. R., Mohan, A. *et al.* Development of the Senseiver for efficient field reconstruction from sparse observations. **Nat. Mach. Intell. 5**, 1317–1325 (2023).

[44] Miller, H. J. Tobler's first law and spatial analysis. **Ann. Assoc. Am. Geogr. 94**, 284–289 (2004).

[45] Zhou, H., Huang, W., Chen, Y. *et al.* Road network representation learning with the third law of geography. In **Adv. Neural Inf. Process. Syst. 37**, 11789–11813 (2024).

[46] Veličković, P., Cucurull, G., Casanova, A. *et al.* Graph attention networks. arXiv 1710.10903 (2017).

[47] Zamir, S. W., Arora, A., Khan, S. *et al.* Restormer: Efficient transformer for high-resolution image restoration. In **Proc. IEEE/CVF Conf. Comput. Vis. Pattern Recognit.**, 5728–5739 (2022).

[48] Zhang, Z., Tan, A., Lin, L. *et al.* CSACL: A channel spatial attention convolutional LSTM model for short-term sea surface temperature prediction. **IEEE Geosci. Remote Sens. Lett. 21**, 1–5 (2023).

[49] Fan, Y., Wei, C., Wong, J. C. *et al.* Cascaded PhyFormer-SRNet: A physics-informed transformer-based framework with cascaded refinement for spatiotemporal super-resolution of turbulent flows. **Phys. Fluids 37** (2025).

[50] Hersbach, H. *et al.* The ERA5 global reanalysis. **Q. J. R. Meteorol. Soc. 146**, 1999–2049 (2020).

# Appendix A: Evaluation of power mapping based on the power curve of a 5 MW wind turbine

To further evaluate the practical value of wind speed reconstruction results in real wind power scenarios, this study supplements the appendix with a power-mapping analysis based on a 5 MW wind turbine. Specifically, a standard 5 MW turbine is adopted, and it is assumed that one turbine is installed at each grid point in the study region. The wind fields reconstructed by different models are then mapped into the corresponding power fields. **Appendix Figure 1** presents the theoretical power curve used in this study: when the wind speed is lower than the cut-in wind speed $v_{in} = 3$m/s, the turbine does not generate power; within the range of 3–11.4 m/s, the power increases approximately cubically with wind speed; when the wind speed reaches the rated wind speed $v_{rated} = 11.4$m/s, the power remains constant at 5 MW; and when the wind speed exceeds the cut-out wind speed $v_{out} = 25$ m/s, the turbine shuts down. It can thus be seen that wind speed errors are not linearly transferred to power errors. Near the cut-in wind speed and within the ramp-up region below the rated wind speed, even small wind speed deviations may lead to more pronounced power estimation errors. Therefore, it is necessary to further evaluate the practical performance of different models from the perspective of power estimation.

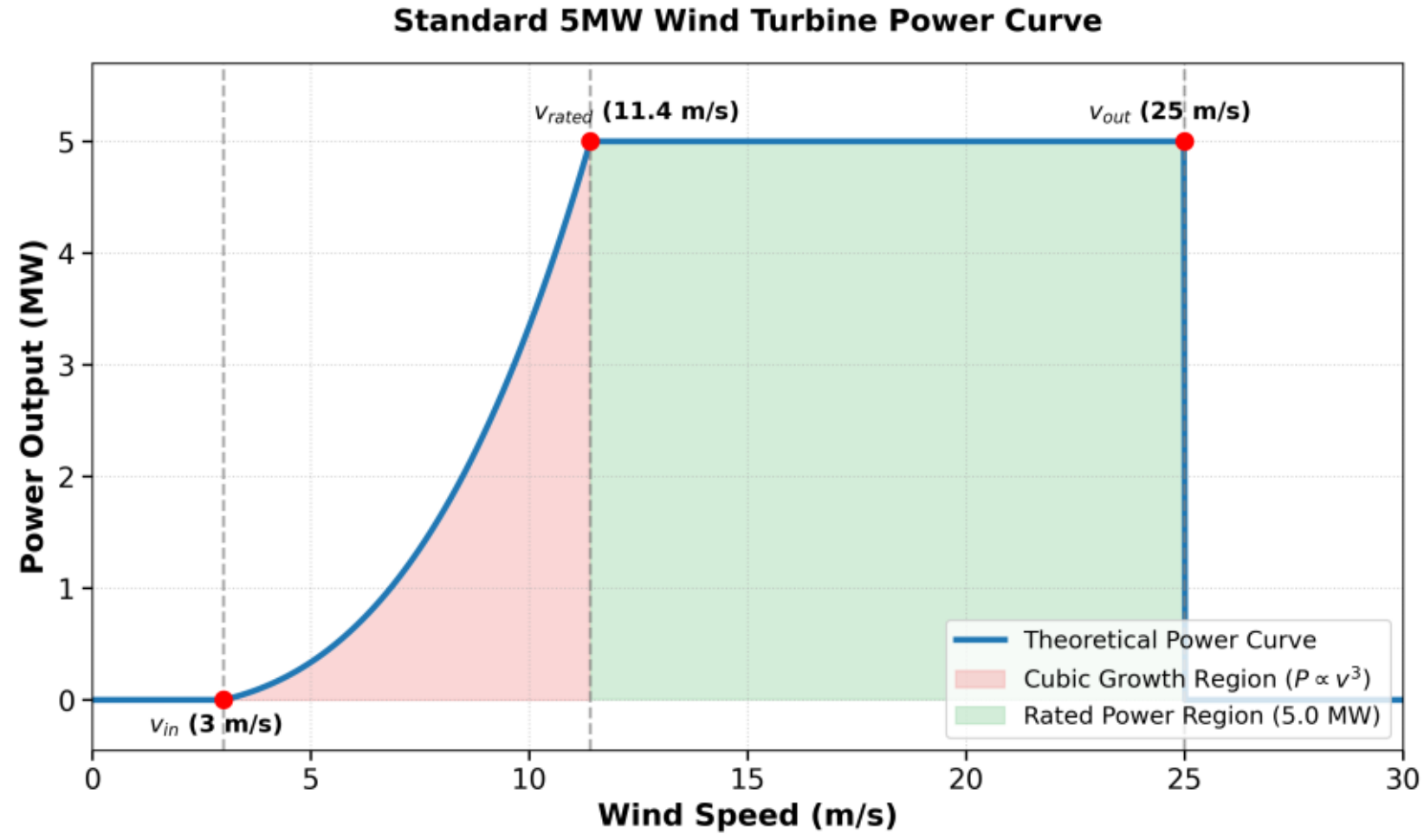

**Appendix Figure 1.** Power curve of the 5 MW wind turbine.

From the overall power error results in **Appendix Table 1**, Zhinv achieves the best performance across all metrics. Its RMSE and MAE are 0.14 MW and 0.064 MW, respectively, both of which are the lowest among all methods, indicating that the power field mapped from its reconstructed wind field is the closest to the reference. In terms of regional total power bias, Zhinv yields an ANE of -44.98 MW, corresponding to a Net% of -1.31%, suggesting that the model exhibits only a minor systematic underestimation and that the bias level is significantly lower than that of Kriging (-358.41 MW, -10.46%), KCN (-141.01 MW, -4.12%), as well as CNN, PConv, and other methods. Furthermore, in terms of the absolute error of regional total power, Zhinv achieves an AAE of 228.61 MW and an Abs% of 6.67%, which are also the lowest among all methods, outperforming the strongest baseline, Kriging+CNN, which records 267.99 MW and 7.82%, respectively. These results indicate that the advantage of Zhinv lies not only in the accuracy of single-grid-point power estimation, but also in the overall accuracy of regional total power statistics. In contrast, traditional Kriging performs significantly worse across all power-related metrics, suggesting that relying solely on spatial interpolation is insufficient to support reliable regional power estimation. Although CNN, PConv, KCN, and Kriging+CNN show clear improvements over traditional methods, they remain inferior to Zhinv in both overall accuracy and total power control capability.

**Appendix Table 1.** Comparison of regional power field estimation errors and total power biases of different methods under the 512-observation-point setting. RMSE(MW) and MAE(MW) denote the root mean square error and mean absolute error of grid-level power estimation, respectively. ANE(MW) and Net% denote the signed bias and relative signed bias of regional total power, respectively. AAE (MW) and Abs% denote the absolute error and relative absolute error of regional total power, respectively.

| Model | RMSE(MW) | MAE(MW) | ANE(MW) | Net%(%) | AAE(MW) | Abs%(%) |
|---|---|---|---|---|---|---|
| Kriging | 0.421 | 0.206 | -358.41 | -10.46 | 737.82 | 21.53 |
| CNN | 0.17 | 0.083 | -108.4 | -3.16 | 297.28 | 8.68 |
| PConv | 0.173 | 0.084 | -107.77 | -3.15 | 300.43 | 8.77 |
| KCN | 0.261 | 0.134 | -141.01 | -4.12 | 479.97 | 14.01 |
| Kriging+CNN | 0.158 | 0.075 | -23.04 | -0.67 | 267.99 | 7.82 |
| ZHINV | 0.14 | 0.064 | -44.98 | -1.31 | 228.61 | 6.67 |

**Appendix Table 2.** Comparison of total absolute error (TAE) of regional total power under extreme scenarios for different methods with 512 observation points. Here, 99% Worst-Case TAE denotes the average TAE over the worst 1% samples ranked by regional total power error; High Variance TAE (Top 10%) denotes the average TAE over the top 10% samples with the strongest spatial variance in the reference wind field; and Peak Load TAE (Top 10%) denotes the average TAE over the top 10% samples with the highest reference regional total power.

| Model | 99% Worst-Case TAE (MW) | High Variance TAE (Top10%) (MW) | Peak Load TAE (Top10%) (MW) |
|---|---|---|---|
| Kriging | 1675.54 | 899.63 | 1102.68 |
| CNN | 616.96 | 389.10 | 473.24 |
| PConv | 618.37 | 391.03 | 477.92 |
| KCN | 937.68 | 637.42 | 775.58 |
| Kriging+CNN | 564.22 | 343.45 | 419.68 |
| ZHINV | 492.16 | 299.39 | 364.54 |

On this basis, we further conducted an extreme-load scenario analysis using the existing samples. Specifically, 99% Worst-Case TAE is used to measure the error level under the worst cases in the regional total power error distribution; High Variance TAE (Top 10%) denotes the total power error for samples with the most intense spatial fluctuations; and Peak Load TAE (Top 10%) denotes the error for samples with the highest regional total power. The results in **Appendix Table 2** show that Zhinv achieves the best performance on all three tail-risk metrics, with 99% Worst-Case TAE, High Variance TAE, and Peak Load TAE of 492.16 MW, 299.39 MW, and 364.54 MW, respectively, all significantly lower than those of the other methods. In comparison, the strongest baseline, Kriging+CNN, records 564.22 MW, 343.45 MW, and 419.68 MW on the corresponding metrics, while traditional Kriging reaches 1675.54 MW, 899.63 MW, and 1102.68 MW, respectively. These results indicate that Zhinv not only reconstructs regional power distributions more accurately under normal conditions, but also maintains stronger stability under complex fluctuation scenarios and peak-load conditions, effectively suppressing the amplification of wind speed errors into regional total power errors.

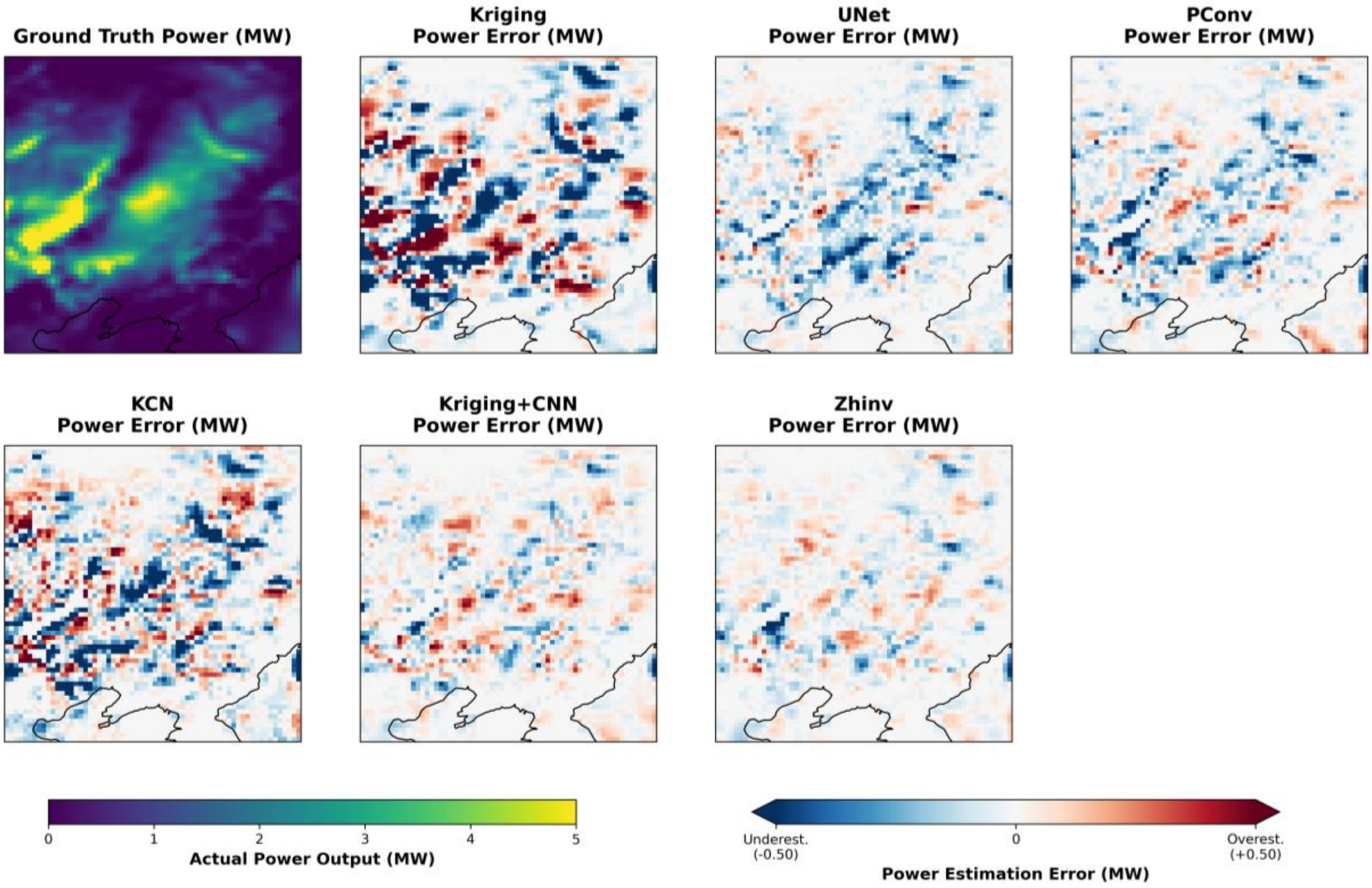

**Appendix Figure 2.** Spatial distribution of 5 MW turbine power estimation errors of different methods under a density of 512 sparse observation points. The upper-left panel shows the ground-truth power distribution, while the subsequent panels present the spatial distributions of the power estimation errors of different methods relative to the truth.

**Appendix Figure 2** further validates the power estimation capability of different methods from the perspective of spatial distribution. Consistent with the quantitative results in **Appendix Table 1** and **Appendix Table 2**, the error maps of Kriging and KCN still exhibit many high-magnitude error patches with alternating positive and negative deviations, indicating clear deficiencies in recovering the spatial structure of the power field, with pronounced local overestimation and underestimation. This is also consistent with their relatively high RMSE, MAE, and regional total power errors. In contrast, the error distributions of UNet, PConv, and Kriging+CNN are significantly more constrained, with both the extent and intensity of large-error regions shrinking, suggesting that deep learning models can improve the reconstruction quality of the power field to a certain extent. Zhinv shows the smoothest error distribution and the smallest error magnitude. In particular, in regions with higher reference power and more pronounced spatial gradients, no obvious large-area systematic overestimation or underestimation is observed, which is consistent with its lowest RMSE, MAE, AvgAbsErr (TAE), and extreme-load scenario errors reported in the table. Overall, this figure, together with the statistical results, demonstrates that Zhinv not only reduces grid-level power estimation errors, but also more effectively suppresses the spatial propagation of errors, thereby maintaining higher reliability in both regional total power estimation and peak-load scenarios.

The results in the Figures and Tables show that nonlinear characteristics of the 5 MW wind turbine power curve amplify the impact of wind speed reconstruction variations on downstream power estimation. Despite this, Zhinv consistently maintains optimal performance in terms of overall power error, regional total bias, and extreme-load scenarios. This indicates that the advancements of the proposed method are not confined to the wind speed field itself but effectively propagate to practical wind power assessment tasks. Consequently, Zhinv provides a more reliable foundation for regional wind power monitoring, total estimation, and operational decision-making.